\documentclass{article}

    \PassOptionsToPackage{numbers, compress}{natbib}

    \usepackage[final]{neurips_data_2021}

\usepackage[utf8]{inputenc} %
\usepackage[T1]{fontenc}    %
\usepackage[hyphens]{url}   %
\usepackage[colorlinks=true,citecolor=gray,urlcolor=gray,linkcolor=blue]{hyperref}
\usepackage{amsfonts}       %
\usepackage{amsmath}
\usepackage{booktabs}       %
\usepackage{enumitem}       %
\usepackage{microtype}      %
\usepackage{nicefrac}       %
\usepackage{graphicx}       %
\usepackage{xcolor}         %
\usepackage[final]{changes}    %
\usepackage{adjustbox}
\usepackage{cleveref}
\usepackage{multirow}
\usepackage{array}
\usepackage{subcaption}

\newcommand{\bench}{\textsc{SustainBench}}
\newcommand{\wilds}{\textsc{Wilds}}

\title{\textsc{SustainBench}: Benchmarks for Monitoring the Sustainable Development Goals with Machine Learning}

\author{%
  Christopher Yeh\thanks{Joint first authors.} \\
  Caltech \\
  \And
  Chenlin Meng$^{*}$ \\
  Stanford \\
  \And
  Sherrie Wang$^{*}$ \\
  UC Berkeley \\
  \And
  Anne Driscoll\thanks{Joint second authors.} \\
  Stanford \\
  \And
  Erik Rozi$^{\dagger}$\\
  Stanford \\
  \AND
  Patrick Liu$^{\dagger}$\\
  Stanford \\
  \And
  Jihyeon Lee$^{\dagger}$\\
  Stanford \\
  \And
  Marshall Burke \\
  Stanford \\
  \And
  David Lobell \\
  Stanford \\
  \And
  Stefano Ermon \\
  Stanford \\
}

\begin{document}

\maketitle

\begin{abstract}
Progress toward the United Nations Sustainable Development Goals (SDGs) has been hindered by a lack of data on key environmental and socioeconomic indicators, which historically have come from ground surveys with sparse temporal and spatial coverage.
Recent advances in machine learning have made it possible to utilize abundant, frequently-updated, and globally available data, such as from satellites or social media, to provide insights into progress toward SDGs.
Despite promising early results, approaches to using such data for SDG measurement thus far have largely evaluated on different datasets or used inconsistent evaluation metrics, making it hard to understand whether performance is improving and where additional research would be most fruitful.
Furthermore, processing satellite and ground survey data requires domain knowledge that many in the machine learning community lack.
In this paper, we introduce \bench, a collection of 15 benchmark tasks across 7 SDGs, including tasks related to economic development, agriculture, health, education, water and sanitation, climate action, and life on land. Datasets for 11 of the 15 tasks are released publicly for the first time.
Our goals for \bench{} are to (1) lower the barriers to entry for the machine learning community to contribute to measuring and achieving the SDGs; (2) provide standard benchmarks for evaluating machine learning models on tasks across a variety of SDGs; and (3) encourage the development of novel machine learning methods where improved model performance facilitates progress towards the SDGs.

\end{abstract}
\section{Introduction}

In 2015, the United Nations (UN) proposed 17 Sustainable Development Goals (SDGs) to be achieved by 2030, for promoting prosperity while protecting the planet~\cite{2015transforming}. The SDGs span social, economic, and environmental spheres, ranging from ending poverty to achieving gender equality to combating climate change (see \Cref{app:tab:sdgs_full}). Progress toward SDGs is traditionally monitored through statistics collected by civil registrations, population-based surveys and censuses.  However, such data collection is expensive and requires adequate statistical capacity, and many countries go decades between making ground measurements on key SDG indicators~\cite{burke2021using}. Only roughly half of SDG indicators have regular data from more than half of the world's countries \cite{un2021tier}. These data gaps severely limit the ability of the international community to track progress toward the SDGs.

\begin{figure}[t]
\centering
\includegraphics[width=\linewidth]{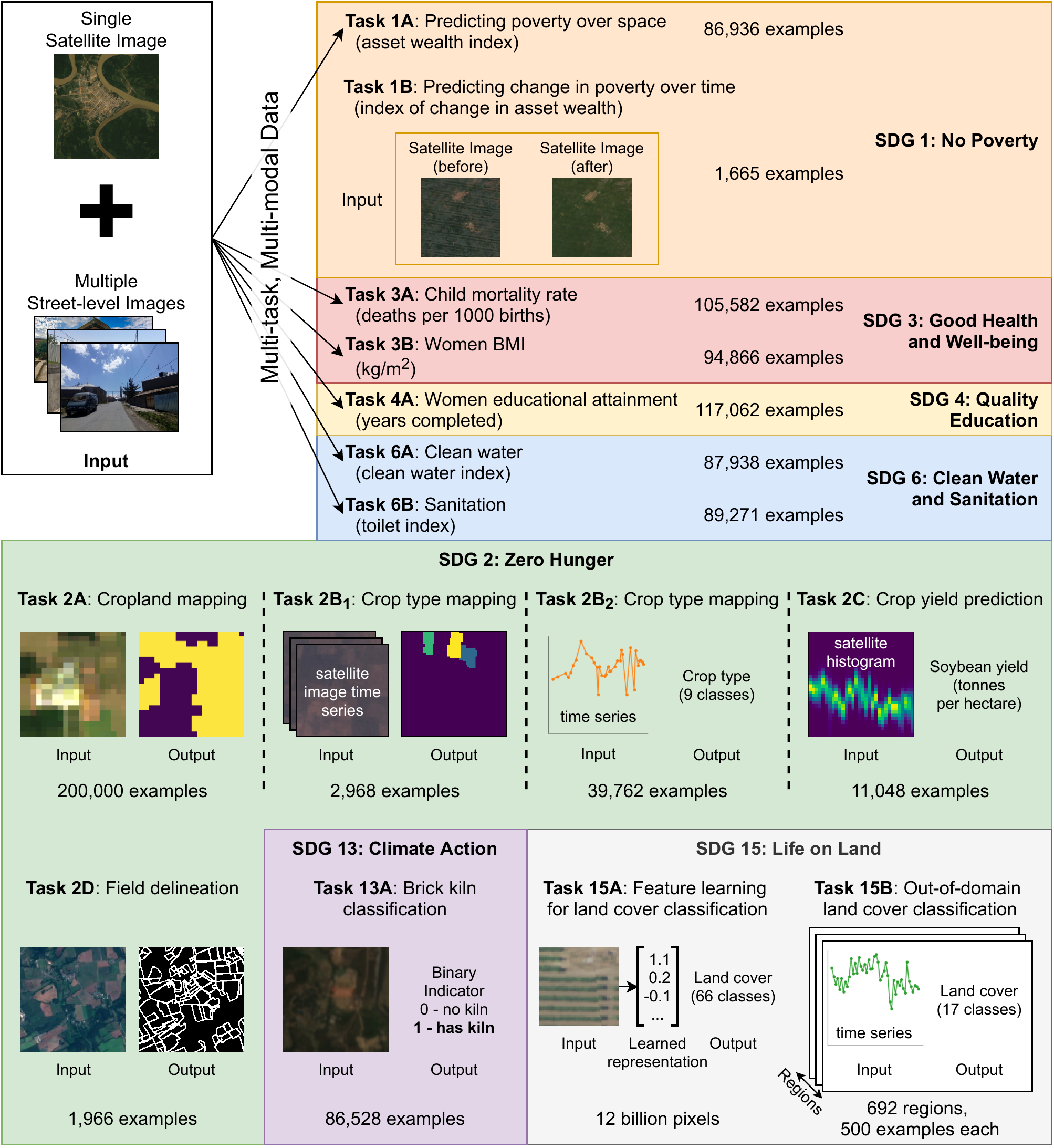}
\caption{Datasets and tasks included in \bench ranging from poverty prediction to land cover classification (described in \Cref{sec:data} with additional details in \Cref{app:sec:dataset}). \emph{Data for 11 out of 15 tasks are publicly released for the first time}.}
\label{fig:datasets_summary}
\vspace*{-0.5cm}
\end{figure}

Advances in machine learning (ML) have shown promise in helping plug these data gaps, demonstrating how sparse ground data can be combined with  abundant, cheap and frequently updated sources of novel sensor data to measure a range of SDG-related outcomes \cite{gsdr2014,burke2021using}. For instance, data from satellite imagery, social media posts, and/or mobile phone activity can predict poverty \cite{blumenstock2015predicting,jean2016combining,yeh2020using}, annual land cover \cite{friedl2002global,buchhorn2020copernicus}, deforestation \cite{hansen2013high,irvin2020forestnet}, agricultural cropping patterns \cite{cdl,wang2020mapping}, crop yields \cite{azzari2017towards, you2017deep}, and the location and impact of natural disasters \cite{debruijin2019a,tellman2021satellite}. As a timely example of real-world impact, the governments of Bangladesh, Mozambique, Nigeria, Togo, and Uganda used ML-based poverty and cropland maps generated from satellite imagery or phone records to target economic aid to their most vulnerable populations during the COVID-19 pandemic \cite{blumenstock2020machine,gentilini2021cash,kerner2020rapid,lowe2021national}. Other recent work demonstrates using ML-based poverty maps to measure the effectiveness of large-scale infrastructure investments~\cite{ratledge2021using}.

But further methodological progress on the ``big data approach'' to monitoring SDGs is hindered by a number of key challenges. First, downloading and working with both novel input data (\emph{e.g.}, from satellites) and ground-based household surveys requires domain knowledge that many in the ML community lack. Second, existing approaches have been evaluated on different datasets, data splits, or evaluation metrics, making it hard to understand whether performance is improving and where additional research would be most fruitful~\cite{burke2021using}. This is in stark contrast to canonical ML datasets like MNIST, CIFAR-10~\cite{krizhevsky2009learning}, and ImageNet~\cite{ILSVRC15} that have standardized inputs, outputs, and evaluation criteria and have therefore facilitated remarkable algorithmic advances~\cite{he2016deep,dinh2016density,kingma2013auto,he2020momentum,huang2017densely}. Third, methods used so far are often adapted from methods originally designed for canonical deep learning datasets (\emph{e.g.}, ImageNet). However, the datasets and tasks relevant to SDGs are unique enough to merit their own methodology. For example, gaps in monitoring SDGs are widest in low-income countries, where only sparse ground labels are available to train or validate predictive models.

To facilitate methodological progress, this paper presents \bench, a compilation of datasets and benchmarks for monitoring the SDGs with machine learning.
Our goals are to
\begin{enumerate}[nosep]
    \item lower the barriers to entry by supplying high-quality domain-specific datasets in development economics and environmental science,
    \item provide benchmarks to standardize evaluation on tasks related to SDG monitoring, and
    \item encourage the ML community to evaluate and develop novel methods on problems of global significance where improved model performance facilitates progress towards SDGs.
\end{enumerate}

In \bench, we curate a suite of 15 benchmark tasks across 7 SDGs where we have relatively high-quality ground truth labels: No Poverty (SDG 1), Zero Hunger (SDG 2), Good Health and Well-being (SDG 3), Quality Education (SDG 4), Clean Water and Sanitation (SDG 6), Climate Action (SDG 13), and Life on Land (SDG 15). \Cref{fig:datasets_summary} summarizes the datasets in \bench. Although results for some tasks have been published previously, \emph{data for 11 of the 15 tasks are being made public for the first time}. We provide baseline models for each task and a public leaderboard\footnote{\url{https://sustainlab-group.github.io/sustainbench/leaderboard}}.

To our knowledge, this is the first set of large-scale cross-domain datasets targeted at SDG monitoring compiled with standardized data splits to enable benchmarking. \bench{} is not only valuable to improving sustainability measurements but also offers tasks for ML challenges, allowing for the development of self-supervised learning (\Cref{sec:sdg15}), meta-learning (\Cref{sec:sdg15}), and multi-modal/multi-task learning methods (\Cref{sec:sdg1,sec:sdg3,sec:sdg4,sec:sdg6}) on real-world datasets.

In the remainder of this paper, \Cref{sec:related} surveys related datasets; \Cref{sec:data} introduces the SDGs and datasets covered by \bench; \Cref{sec:benchmarks} summarizes state-of-the-art models on each dataset and where methodological advances are needed; and \Cref{sec:discussion} highlights the impact, limitations, and future directions of this work. The \hyperref[sec:app]{Appendix} includes detailed information about the inputs, labels, and tasks for each dataset.

\section{Related Work}
\label{sec:related}

Our work builds on a growing body of research that seeks to measure SDG-relevant indicators, including those cited above. These individual studies typically focus on only one SDG-related task, but even within a specific SDG domain (\emph{e.g.}, poverty prediction), most tasks lack standardized datasets with clear replicate-able benchmarks~\cite{burke2021using}. In comparison, \bench{} is a compilation of datasets that covers 7 SDGs and provides 15 standardized, replicate-able tasks with established benchmarks. \Cref{tab:compare_benchmarks} compares \bench{} against existing datasets that pertain to SDGs, are publicly available, provide ML-friendly inputs/outputs, and specify standardized evaluation metrics.

Perhaps the most closely-related benchmark dataset is \wilds{}~\cite{koh2021wilds}, which provides a comprehensive benchmark for distribution shifts in real-world applications. However, \wilds{} is not focused on SDGs, and although it includes a poverty mapping task, our poverty dataset covers 5$\times$ more countries.

There also exist a number of datasets for performing satellite or aerial imagery tasks related to the SDGs~\cite{christie2018functional,schmitt2019sen12ms,sumbul2019bigearthnet,yang2010bag,vanetten2019spacenet,lam2018xView,gupta2019xbd,diux2021xview3,demir2018deepglobe,vanetten2019spacenet} which share similarities with the inputs of \bench{} on certain benchmarks. For example, \cite{schmitt2019sen12ms} compiled imagery from the Sentinel-1/2 satellites, which we also use for SDG monitoring tasks, and the Radiant Earth Foundation has compiled datasets for crop type mapping \cite{radiant}, a task we also include. However, \bench{}'s goal is to provide a broader view of what ML can do for SDG monitoring; it is differentiated in its focus on multiple SDGs, multiple inputs, and on low-income regions in particular. For tasks where existing datasets are abundant (\emph{e.g.}, cropland and land cover classification), \bench{} has tasks that address remaining challenges in the domain (\emph{e.g.}, learning from weak labels, sharing knowledge across the globe). \Cref{app:sec:dataset} provides task-by-task comparisons of \bench{} datasets with prior work.

\begin{table}[tbp]
\centering
\caption{A comparison of \bench{} with related datasets and benchmarks. A dataset is only included if it is relevant for an SDG, is publicly available, provides both inputs and outputs in ML-friendly formats, defines train/test sets, and standardizes evaluation metrics.}
\label{tab:compare_benchmarks}
\begin{adjustbox}{max width=\linewidth}
\begin{tabular}{>{\raggedright\arraybackslash}p{4.0cm} >{\raggedright\arraybackslash}p{3cm} >{\raggedright\arraybackslash}p{2cm} >{\raggedright\arraybackslash}p{2cm} >{\raggedright\arraybackslash}p{2.5cm}
c c c c c c c c c c c}
\toprule
& & & & & \multicolumn{9}{c}{\textbf{Relevant for SDGs}} \\
\cmidrule{6-14}
\textbf{Name}
    & \textbf{Purpose}
    & \textbf{Geography}
    & \textbf{Time}
    & \textbf{Inputs}
    & \textbf{1}
    & \textbf{2}
    & \textbf{3}
    & \textbf{4}
    & \textbf{6}
    & \textbf{11}
    & \textbf{13}
    & \textbf{14}
    & \textbf{15}
    \\
\midrule
\bench{}
    & SDG monitoring
    & 1-105 countries/task \newline (119 total)
    & 1-24 years/task \newline in 1996-2019
    & Sat. images, street-level images, and/or time series
    & \checkmark & \checkmark & \checkmark & \checkmark & \checkmark & & \checkmark & & \checkmark
    \\
    \midrule
Yeh \textit{et al.} / \wilds{} \cite{yeh2020using,koh2021wilds}
    & Poverty mapping
    & 23 countries
    & 2009-16
    & Sat. images
    & \checkmark
    \\
    \midrule
Radiant MLHub \cite{radiant}
    & Crop type mapping
    & 8 countries
    & 1-3 years/task \newline in 2015-21
    & Sat. time series or drone images
    & & \checkmark & & & & & & &
    \\
    \midrule
SpaceNet \cite{vanetten2019spacenet}
    & Building \& road detection
    & 10+ cities
    & Unknown
    & Sat. images \& time series
    & & & & & & \checkmark & & &
    \\
    \midrule
DeepGlobe \cite{demir2018deepglobe}
    & Building \& road detection,\newline land cover classification
    & 3 countries, \newline 4 cities
    & Unknown
    & Sat. images
    & & & & & & \checkmark & & & \checkmark
    \\
    \midrule
fMoW / \wilds{} \cite{christie2018functional, koh2021wilds}
    & Object detection
    & 207 countries
    & 2002-17
    & Sat. images
    & & & & & & \checkmark & & &
    \\
    \midrule
xView \cite{lam2018xView}
    & Object classification
    & 30+ countries
    & Unknown
    & Sat. images
    & & & & & & \checkmark & & &
    \\
    \midrule
xBD (xView2) \cite{gupta2019xbd}
    & Disaster damage assessment
    & 10 countries
    & 2011-19
    & Sat. images
    & & & & & & \checkmark
    \\
    \midrule
xView3 \cite{diux2021xview3}
    & Illegal fishing detection
    & Oceans
    & Unknown
    & Sat. images
    & & & & & & & & \checkmark
    \\
    \midrule
BigEarthNet \cite{sumbul2019bigearthnet}
    & Land cover classification
    & 10 countries in Europe
    & 2017-18
    & Sat. images
    & & & & & & & & & \checkmark
    \\
    \midrule
ForestNet \cite{irvin2020forestnet}
    & Deforestation drivers
    & Indonesia
    & 2001-16
    & Environ. data \& sat. images
    & & & & & & & \checkmark & & \checkmark
    \\
    \midrule
iWildCam2020 /\newline \wilds{} \cite{beery2020iwildcam,koh2021wilds}
    & Wildlife monitoring
    & 12 countries
    & 2013-15
    & Camera trap images
    & & & & & & & & & \checkmark
    \\
\bottomrule
\end{tabular}
\end{adjustbox}
\vspace*{-0.3cm}
\end{table}
\section{\bench{} Datasets and Tasks}
\label{sec:data}

\begin{figure}[bp]
    \centering
    \includegraphics[width=\linewidth]{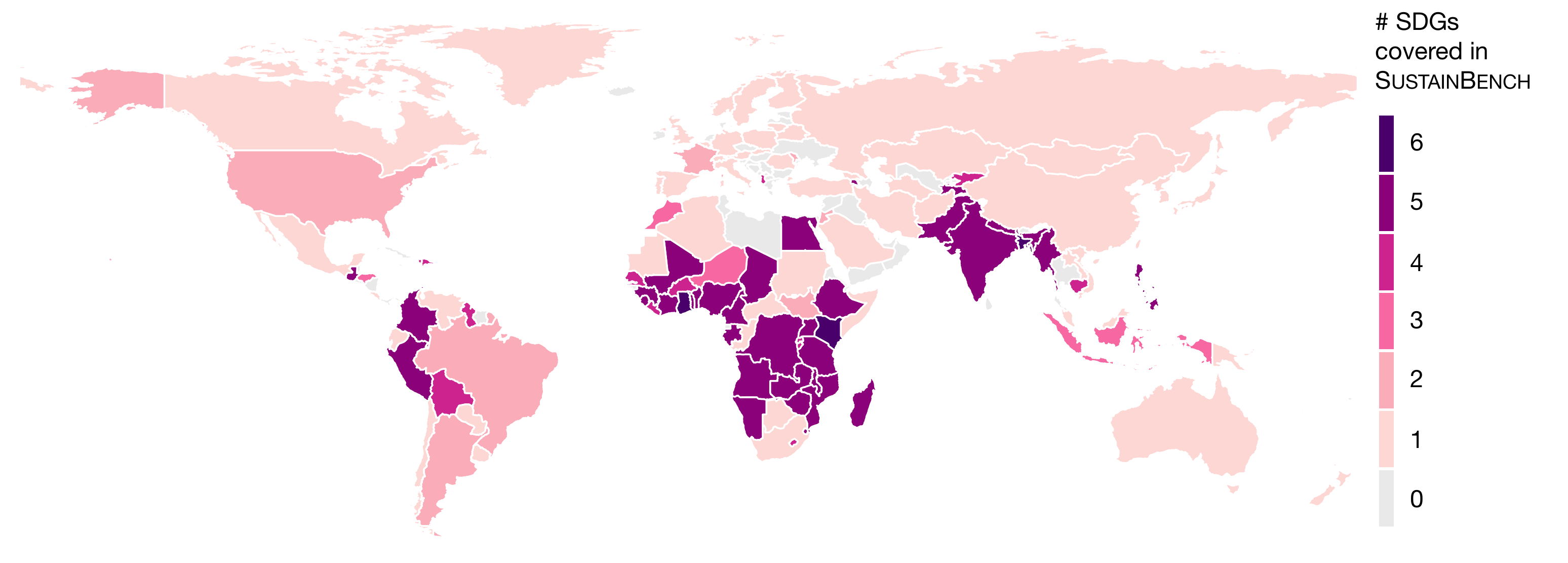}
    \caption{A map of how many SDGs are covered in \bench{} for every country. \bench{} has global coverage with an emphasis on low-income countries. In total, 119 countries have at least one task in \bench.}
    \label{fig:map}
    \vspace*{-0.5cm}
\end{figure}

In this section, we introduce the \bench{} datasets and provide background on the SDGs that they help monitor. Seven SDGs are currently covered: No Poverty (SDG 1), Zero Hunger (SDG 2), Good Health and Well-being (SDG 3), Quality Education (SDG 4), Clean Water and Sanitation (SDG 6), Climate Action (SDG 13), and Life on Land (SDG 15). We describe how progress toward each goal is traditionally monitored, the gaps that currently exist in monitoring, and how certain indicators can be monitored using non-traditional datasets instead. \Cref{fig:datasets_summary} summarizes the SDG, inputs, outputs, tasks, and original reference of each dataset, and \Cref{fig:map,app:fig:sdgmaps} visualize how many SDG indicators are covered by \bench{} in each country. All of the datasets are easily downloaded via a Python package that integrates with the PyTorch ML framework~\cite{paszke2019pytorch}.

\subsection{No Poverty (SDG 1)}
\label{sec:sdg1}

Despite decades of declining poverty rates, an estimated 8.4\% of the global population remains in extreme poverty as of 2019, and progress has slowed in recent years~\cite{un2021sdgs}. But data on poverty remain surprisingly sparse, hampering efforts at monitoring local progress, targeting aid to those who need it, and evaluating the effectiveness of antipoverty programs \cite{burke2021using}. In most African countries, for example, nationally representative consumption or asset wealth surveys, the key source of internationally comparable poverty measurements, are only available once every four years or less~\cite{yeh2020using}.

For \bench, we processed survey data from two international household survey programs: Demographic and Health Surveys (DHS) \cite{icf1996dhs} and the Living Standards Measurement Study (LSMS). Both constitute nationally representative household-level data on assets, housing conditions, and education levels, among other attributes. Notably, only LSMS data form a panel---\emph{i.e.}, the same households are surveyed over time, facilitating comparison over time. Using a a principal components analysis (PCA) approach~\cite{filmer2001estimating,sahn2003exploring}, we summarize the survey data into a single scalar asset wealth index per ``cluster,'' which roughly corresponds to a village or local community. We refer to cluster-level wealth (or its absence) as ``poverty''. Previous research has shown that widely-available imagery sources including satellite imagery~\cite{jean2016combining,yeh2020using} and crowd-sourced street-level imagery~\cite{lee2021predicting} can be effective for predicting cluster-level asset wealth when used as inputs in deep learning models.

\bench{} includes two regression tasks for poverty prediction at the cluster level, both using imagery inputs to estimate an asset wealth index. The first task (\Cref{sec:sdg1:over_space}) predicts poverty over space, and the second task (\Cref{sec:sdg1:change}) predicts poverty changes over time.

\subsubsection{Poverty Prediction Over Space}
\label{sec:sdg1:over_space}
The poverty prediction over space task involves predicting a cluster-level asset wealth index which represents the ``static'' asset wealth of a cluster at a given point in time. For this task, the labels and inputs are created in a similar manner as in~\cite{yeh2020using}, but with about 5$\times$ as many examples.

\textbf{Dataset}\quad
Following techniques developed in previous works~\cite{jean2016combining,yeh2020using}, we assembled asset wealth data for 2,079,036 households living in 86,936 clusters across 48 countries, drawn from DHS surveys conducted between 1996 and 2019, computing a cluster-level asset wealth index as described above. We provide satellite and street-level imagery inputs, gathered and processed according to established procedures~\cite{yeh2020using,lee2021predicting}. The 255$\times$255$\times$8px satellite images have 7 multispectral bands from Landsat daytime satellites and 1 nightlights band from either the DMSP or VIIRS satellites. The images are rescaled to a resolution of 30m/px and are geographically centered around each surveyed cluster's geocoordinates. Geocoordinates in the public survey data are ``jittered'' by up to 10km from the true locations to protect the privacy of surveyed households \cite{burgert2013geographic}. For each cluster location, we also retrieved up to 300 crowd-sourced, street-level imagery from Mapillary. We evaluate model performance using the squared Pearson correlation coefficient ($r^2$) between predicted and observed values of the asset wealth index on held-out test countries. \Cref{app:dhs} has more dataset details.

\subsubsection{Poverty Prediction Over Time}
\label{sec:sdg1:change}

For predicting temporal changes in poverty, we construct a PCA-based index of changes in asset ownership using LSMS data. For this task, the labels and inputs provided are similar to \cite{yeh2020using}, with small improvements in image and label quality.

\textbf{Dataset}\quad
We provide labels for 1,665 instances of cluster-level asset wealth change from 1,287 clusters in 5 African countries. We use the same satellite imagery sources from the previous poverty prediction task. In this task, however, for each cluster we provide images from the two points in time (before and after) used to compute the difference in asset ownership, instead of only from a single point in time. Because street-level images were only available for $\sim$1\% of clusters, we do not provide them for this task. We evaluate model performance using the squared Pearson correlation coefficient ($r^2$) on predictions and labels in held-out cluster locations. \Cref{app:lsms} has more dataset details.

\subsection{Zero Hunger (SDG 2)}
\label{sec:sdg2}
The number of people who suffer from hunger has risen since 2015, with 690 million or 9\% of the world's population affected by chronic hunger \cite{un2021sdgs}. At the same time, 40\% of habitable land on Earth is already devoted to agricultural activities, making agriculture by far the largest human impact on the natural landscape \cite{fao2021food}.
The second SDG is to ``end hunger, achieve food security and improved nutrition, and promote sustainable agriculture.''
In addition to ending hunger and malnutrition in all forms, the targets under SDG 2 include doubling the productivity of small-scale food producers and promoting sustainable food production \cite{un2021sdgs}.
While traditionally data on agricultural practices and farm productivity are obtained via farm surveys, such data are rare and often of low quality \cite{burke2021using}. Satellite imagery offers the opportunity to monitor agriculture more cheaply and more accurately, by mapping cropland, crop types, crop yields, field boundaries, and agricultural practices like cover cropping and conservation tillage. We discuss the \bench{} datasets for SDG 2 below.

\subsubsection{Cropland mapping with weak labels}

One indicator for SDG 2 is the proportion of agricultural area under productive and sustainable agriculture \cite{un2021sdgs}. Existing state-of-the-art datasets on land cover \cite{buchhorn2020copernicus,friedl2002global} are derived from satellite time series and include a cropland class. However, the maps are known to have large errors in regions of the world like Sub-Saharan Africa where ground labels are sparse \cite{kerner2020rapid}. Therefore, while mapping cropland is largely a solved problem in settings with ample labels, devising methods to efficiently generate georeferenced labels and accurately map cropland in low-resource regions remains an important and challenging research direction.

\textbf{Dataset}\quad
We release a dataset for performing weakly supervised cropland classification in the U.S. using data from \cite{wang2020weakly}, which has not been released previously. While densely segmented labels are time-consuming and infeasible to generate for a large region like Africa, pixel-level and image-level labels are easier to create. The inputs are image tiles taken by the Landsat satellites and composited over the 2017 growing season, and the labels are either binary $\{\text{cropland}, \text{not cropland} \}$ at single pixels or $\{ \ge 50\% \text{ cropland}, < 50\% \text{ cropland}\}$ for the entire image. Labels are generated from a high-quality USDA dataset on land cover \cite{cdl}. Train, validation, and test sets are split along geographic blocks, and we evaluate models by overall accuracy and F1-score. We also encourage the use of semi-supervised and active learning methods to relieve the labeling burden needed to map cropland.

\subsubsection{Crop type mapping in Sub-Saharan Africa}
\label{sec:sdg2:croptype}

Spatially disaggregated crop type maps are needed to assess agricultural diversity and estimate yields. In high-income countries across North America and Europe, crop type maps are produced annually by departments of agriculture using farm surveys and satellite imagery \cite{cdl}. However, no such maps are regularly available for middle- and low-income countries.
Mapping crop types in the Global South faces challenges of irregularly shaped fields, small fields, intercropping, sparse ground truth labels, and highly heterogeneous landscapes
\cite{rustowicz2019semantic}. %
We release two crop type datasets in Sub-Saharan Africa and point the reader to additional datasets hosted by the Radiant Earth Foundation \cite{radiant} (\Cref{tab:compare_benchmarks}). We recommend that ML researchers use all available datasets to ensure model generalizability.

\textbf{Dataset \#1}\quad
We re-release the dataset from \cite{rustowicz2019semantic} in Ghana and South Sudan in a format more familiar to the ML community. The inputs are growing season time series of imagery from three satellites (Sentinel-1, Sentinel-2, and PlanetScope) in 2016 and 2017, and the outputs are semantic segmentation of crop types. Ghana samples are labeled for maize, groundnut, rice, and soybean, while South Sudan samples are labeled for maize, groundnut, rice, and sorghum. We use the same train, validation, and test sets as \cite{rustowicz2019semantic}, which preserve relative percentages of crop types across the splits. We evaluate models using overall accuracy and macro F1-score.

\textbf{Dataset \#2}\quad
We release the dataset used in \cite{kluger2021two} and \cite{jin2019smallholder} to map crop types in three regions of Kenya. Since the timing of growth and spectral signature are two main ways to distinguish crop types, the inputs are annual time series from the Sentinel-2 multi-spectral satellite. The outputs are crop types (9 possible classes). There are a total of 39,762 pixels belonging to 5,746 fields. The training, validation, and test sets are split along region rather than by field in order to develop models that generalize across geography. Our evaluation metrics are overall accuracy and macro-F1 score.

\subsubsection{Crop yield prediction in North and South America}

In order to double the productivity (or yield) of smallholder farms, we first have to measure it, and accurate local-level yield measurements are exceedingly rare in most of the world. %
In \bench, we release county-level yields collected from various government databases; these can still aid in forecasting production, evaluating agricultural policy, and assessing the effects of climate change.

\textbf{Dataset}\quad \label{modis}
Our dataset is based on the datasets used in
\cite{you2017deep} and \cite{wang2018transfer}. We release county-level yields for 857 counties in the U.S., 135 in Argentina, and 32 in Brazil for the years 2005-16.
The inputs are spectral band and temperature histograms over each county for the harvest season from the MODIS satellite.
The ground truth labels are the regional soybean yield per harvest, in metric tonnes per cultivated hectare, retrieved from government data. See \Cref{app:sec:crop} for more details. Models are evaluated using root mean squared error (RMSE) and $R^2$ of predictions with the ground truth. The imbalance of data by country motivates the use of transfer learning approaches.

\subsubsection{Field delineation in France}

Since agricultural practices are usually implemented on the level of an entire field, field boundaries can help reduce noise and improve performance when mapping crop types and yields. Furthermore, field boundaries are a prerequisite for today's digital agriculture services that help farmers optimize yields and profits \cite{waldner2020deep}. Statistics that can be derived from field delineation, such as the size and distribution of crop fields, have also been used to study productivity \cite{carter1984identification,desiere2018land}, mechanization \cite{kuemmerle2013challenges}, and biodiversity \cite{geiger2010persistent}. Field boundary datasets are rare and only sparsely labeled in low-income regions, so we release a large dataset from France to aid in model development.

\textbf{Dataset}\quad
We re-release the dataset introduced in \citealt{aung2020farm}. The dataset consists of Sentinel-2 satellite imagery in France
over 3 time ranges: January-March, April-June, and July-September in 2017. The image has resolution 224$\times$224 corresponding to a 2.24km$\times$2.24km area on the ground. Each satellite image comes along with the corresponding binary masks of boundaries and areas of farm parcels. The dataset consists of a total of 1966 samples. We use a different data split from \cite{aung2020farm} to remove overlapping between the train, validation and test split. Following ~\cite{aung2020farm}, we use the Dice score between the ground truth boundaries and predicted boundaries as the performance metric.

\subsection{Good Health and Well-being (SDG 3)}
\label{sec:sdg3}

Despite significant progress on improving global health outcomes (\emph{e.g.}, halving child mortality rates since 2000 \cite{un2021sdgs}), the lack of local-level measurements in many developing countries continues to constrain the monitoring, targeting, and evaluation of health interventions. We examine two health indicators: female body mass index (BMI), a key input to understanding both food insecurity and obesity; and child mortality rate (deaths under age 5), an official SDG 3 indicator considered to be a summary measure of a society's health. Previous works have demonstrated using satellite imagery \cite{maharana2018use} or street-level Mapillary imagery inputs \cite{lee2021predicting} for predicting BMI. While we are unaware of any prior works using such imagery inputs for predicting child mortality rates, ``there is evidence that child mortality is connected to environmental factors such as housing quality, slum-like conditions, and neighborhood levels of vegetation'' \cite{jankowska2013estimating}, which are certainly observable in imagery.

\textbf{Dataset}\quad
We provide cluster-level average labels for women's BMI and child mortality rates compiled from DHS surveys. There are 94,866 cluster-level BMI labels computed from 1,781,403 women of childbearing age (15-49), excluding pregnant women. There are 105,582 cluster-level labels for child mortality rates computed from 1,936,904 children under age 5. As in the poverty prediction over space task (\Cref{sec:sdg1:over_space}), the inputs for predicting the health labels are satellite and street-level imagery, and models are evaluated using the $r^2$ metric on labels from held-out test countries.

\subsection{Quality Education (SDG 4)}
\label{sec:sdg4}

SDG 4 includes targets that by 2030, all children and adults ``complete free, equitable and quality primary and secondary education''. Increasing educational attainment (measured by years of schooling completed) is known to increase wealth and social mobility,  and higher educational attainment in women is strongly associated with improved child nutrition and decreased child mortality~\cite{graetz2018mapping}. Previous works have demonstrated the ability of deep learning methods to predict educational attainment from both satellite images \cite{zhao2020framework} and street-level images \cite{gebru2017using,lee2021predicting}.

\textbf{Dataset}\quad
We provide cluster-level average years of educational attainment by women of reproductive age (15-49) compiled from same DHS surveys used for creating the asset wealth labels in the poverty prediction task. The 122,435 cluster-level labels were computed from 3,013,286 women across 56 countries. As in the poverty prediction over space task (\Cref{sec:sdg1:over_space}), the inputs for predicting women educational attainment are satellite and street-level imagery, and models are evaluated using the $r^2$ metric on labels from held-out test countries.

\subsection{Clean Water and Sanitation (SDG 6)}
\label{sec:sdg6}

Clean water and sanitation are fundamental to human health, but as of 2020, two billion people globally do not have access to safe drinking water, and 2.3 billion lack a basic hand-washing facility with soap and water~\cite{sachs2021sustainable}. Access to improved sanitation and clean water is known to be associated with lower rates of child mortality~\cite{deshpande2020mapping,fink2011effect}.

\textbf{Dataset}\quad
We provide cluster-level average years of a water quality index and sanitation index compiled from same DHS surveys used for creating the asset wealth labels in the poverty prediction task. The 87,938 (water index) and 89,271 (sanitation index) cluster-level labels were computed from 2,105,026 (water index) and 2,143,329 (sanitation index) households across 49 countries. As in the poverty prediction over space task (\Cref{sec:sdg1:over_space}), the inputs for predicting the water quality and sanitation indices are satellite and street-level imagery, and models are evaluated using the $r^2$ metric on labels from held-out test countries. Since \bench{} includes labels for child mortality in many of the same clusters with sanitation index labels, we encourage researchers to take advantage of the known associations between these variables.

\subsection{Climate Action (SDG 13)}
SDG 13 aims at combating climate change and its disruptive impacts on national economies and local livelihoods~\cite{martin_climate_nodate}. Monitoring emissions and environmental regulatory compliance are key steps toward SDG 13.

\subsubsection{Brick kiln mapping}

Brick manufacturing is a major source of carbon emissions and air pollution in South Asia, with an industry largely comprised of small-scale, informal producers. Identifying brick kilns from satellite imagery is a scalable method to improve compliance with environmental regulations and measure their impact on nearby populations. A recent study \cite{lee2021scalable} trained a CNN to detect kilns and hand-validated the predictions, providing ground truth kiln locations in Bangladesh from October 2018 to May 2019.

\textbf{Dataset}\quad
The high-resolution satellite imagery used in \cite{lee2021scalable} could not be shared publicly because they were proprietary. Hence, we provide a lower resolution alternative---Sentinel-2 imagery, which is available through Google Earth Engine~\cite{gorelick2017google}. We retrieved $64 \times 64 \times 13$ tiles at 10m/pixel resolution from the same time period and labeled each image as not containing a brick kiln (class 0) or containing a brick kiln (class 1) based on the ground truth locations in \cite{lee2021scalable}. There were 6,329 positive examples out of 374,000 examples total; we sampled 25\% of the negative examples and removed null values, resulting in 67,284 negative examples. More details can be found in \Cref{app:remote_sensing}.

\subsection{Life on Land (SDG 15)}
\label{sec:sdg15}

Human activity has altered over 75\% of the earth's surface, reducing forest cover, degrading once-fertile land, and threatening an estimated 1 million animal and plant species with extinction \cite{un2021sdgs}. Our understanding of land cover---\emph{i.e.}, the physical material on the surface of the earth---and its changes is not uniform across the globe. Existing state-of-the-art land cover maps \cite{buchhorn2020copernicus} are significantly more accurate in high-income regions than low-income ones, as the latter have few ground truth labels \cite{kerner2020rapid}. The following two datasets seek to reduce this gap via representation learning and transfer learning.

\subsubsection{Representation learning for land cover classification}

One approach to increase the performance of land cover classification in regions with few labels is to use unsupervised or self-supervised learning to improve satellite/aerial image representations, so that downstream tasks require fewer labels to perform well.

\textbf{Dataset}\quad
We release the high-resolution aerial imagery dataset from \cite{jean2019tile2vec}, which spans a 2500km$^{2}$ (12 billion pixel) area  of Central Valley, CA in the U.S. The output is image-level land cover (66 classes), where labels are generated from a high-quality USDA dataset \cite{cdl}. The region is divided in geographically-continuous blocks into train, validation, and test sets. The user may use the training imagery in any way to learn representations, and we provide a test set of up to 200,000 tiles (100$\times$100px) for evaluation. The evaluation metrics are overall accuracy and macro F1-score.

\subsubsection{Out-of-domain land cover classification}

A second strategy for increasing performance in label-scarce regions is to transfer knowledge learned from classifying land cover in high-income regions to low-income ones.

\textbf{Dataset}\quad
We release the global dataset of satellite time series from \cite{wang2020meta}. The dataset samples 692 regions of size $10 \text{km} \times 10 \text{km}$ around the globe; for each region, 500 latitude/longitude coordinates are sampled. The input is time series from the MODIS satellite over the course of a year, and the output is land cover type (17 possible classes). Users have the option of splitting regions into train, validation, and test sets at random or by continent. The evaluation metrics are overall accuracy, F1-score, and kappa score. The results from \cite{wang2020meta} are reported with all regions from Africa as the test set, but the user can choose to hold out other continents, for which the label quality will be higher.

\section{Results for Baseline Models}
\label{sec:benchmarks}

\bench provides a benchmark and public leaderboard website for the datasets described in \Cref{sec:data}. Each dataset has standard train-test splits with well-defined performance metrics detailed in \Cref{app:baseline_models}. We also welcome community submissions using additional data sources beyond what is provided in \bench, such as for pre-training or regularization. \Cref{tab:benchmark} summarizes the baseline models and results. Code to reproduce our baseline models is available on GitHub\footnote{\url{https://github.com/sustainlab-group/sustainbench/}}.

Here, we highlight some main takeaways from our baseline models. First, there is significant room for improvement for models that can take advantage of multi-modal inputs. Specifically, our baseline model for the DHS survey-based tasks only uses the satellite imagery inputs, and its poor performance on predicting child mortality and women educational attainment demonstrates the need to leverage additional data sources, such as the street-level imagery we provide. Second, ML model development can lead to significant gains in performance for SDG-related tasks. While the original paper that compiled \bench{}'s field delineation dataset achieved a Dice score of 0.61 with a standard U-Net \cite{aung2020farm}, we applied a new attention-based CNN developed specifically for field delineation \cite{waldner2021detect} and achieved a 0.87 Dice score. For more task-specific discussions, please see \Cref{app:baseline_models}.

\begin{table}
\centering
\caption{\added{Benchmark performance on 15 tasks across 7 SDGs. See details in \Cref{app:baseline_models}. For the Model Type column, kNN = k-nearest neighbors, GP = Gaussian process. An asterisk (*) indicates a result on a similar dataset, but not the exact \bench{} test set.}
}
\label{tab:benchmark}
\begin{adjustbox}{max width=0.99\linewidth}
\begin{tabular}{>{\raggedright\arraybackslash}p{0.18\textwidth} >{\raggedright\arraybackslash}p{0.3\textwidth} c c >{\raggedright\arraybackslash}p{0.2\textwidth} >{\raggedright\arraybackslash}p{0.2\textwidth} c}
\toprule
\textbf{SDG} & \textbf{Task} &
\textbf{Countries} &
\textbf{Metric} & \textbf{Benchmark Value} & \textbf{Model Type} & \textbf{Ref} \\
\midrule
\multirow{2}{*}{No Poverty}
    & Poverty prediction over space
    & 48 countries
    & $r^2$
    & 0.63
    & kNN
    & \cite{yeh2020using} \\
    & Poverty prediction over time
    & 5 African countries
    & $r^2$
    & 0.35*
    & ResNet-18
    & \cite{yeh2020using} \\
\midrule
\multirow{8}{*}{Zero Hunger}
    & Weakly supervised cropland classification
    & \multirow{2}{*}{United States} 
    & \multirow{2}{*}{F1 score} 
    & 0.88 (pixel label) 0.80 (image label)
    & \multirow{2}{*}{U-Net}
    & \multirow{2}{*}{\cite{wang2020weakly}} \\
\cmidrule{2-7}
    & \multirow{2}{*}{Crop type classification}
    & Ghana, South Sudan 
    & Macro F1 
    & 0.57, 0.70 
    & LSTM
    & \cite{rustowicz2019semantic} \\
    & 
    & Kenya
    & Macro F1 
    & 0.30
    & Random forest
    & \cite{kluger2021two} \\
\cmidrule{2-7}
    & \multirow{2}{*}{Crop yield prediction}
    & United States 
    & \multirow{2}{*}{RMSE} 
    & 0.37 t/ha 
    & CNN+GP
    & \cite{you2017deep} \\
    & 
    & Argentina, Brazil 
    &  
    & 0.62 t/ha, 0.42 t/ha 
    & LSTM
    & \cite{wang2018transfer} \\
\cmidrule{2-7}
    & \multirow{2}{*}{Field delineation}
    & \multirow{2}{*}{France}
    & \multirow{2}{*}{Dice score}
    & 0.61  
    & U-Net
    & \cite{aung2020farm} \\
    & 
    &
    &
    & 0.87
    & \added{FracTAL Res-UNet}
    & \cite{waldner2021detect} \\
\midrule
Good Health
    & Child mortality rate
    & 56 countries
    & $r^2$
    & 0.01
    & kNN
    & -- \\
\& Well-Being
    & Women BMI
    & 53 countries
    & $r^2$
    & 0.42
    & kNN
    & -- \\
\midrule
Quality Education
    & Women education
    & 53 countries
    & $r^2$
    & 0.26
    & kNN
    & -- \\
\midrule
Clean Water
    & Water index
    & 49 countries
    & $r^2$
    & 0.40
    & kNN
    & -- \\
and Sanitation
    & Sanitation index
    & 49 countries
    & $r^2$
    & 0.36
    & kNN
    & -- \\
\midrule
Climate Action
    & Brick kiln detection
    & Bangladesh
    & Accuracy
    & 0.94*
    & ResNet-50
    & \cite{lee2021scalable}
    \\
\midrule
\multirow{4}{*}{Life on Land} 
    & Representation learning for land cover 
    & \multirow{2}{*}{United States}
    & \multirow{2}{*}{Accuracy}
    & 0.55 ($n=1,000$) 0.58 ($n=10,000$) 
    & Tile2Vec with ResNet-50
    & \multirow{2}{*}{\cite{jean2019tile2vec}} \\
\cmidrule{2-7}
    & Out-of-domain land cover classification 
    & \multirow{2}{*}{Global}
    & \multirow{2}{*}{Kappa}
    & 0.32 (1-shot, 2-way) 
    & MAML with shallow 1D CNN
    & \multirow{2}{*}{\cite{wang2020meta}} \\
\bottomrule
\end{tabular}
\end{adjustbox}
\vspace*{-0.5cm}
\end{table}

\section{Impact, Limitations, and Future Work}
\label{sec:discussion}

This paper introduces \bench, which, to the best of our knowledge, is the largest compilation to date of datasets and benchmarks for monitoring the SDGs with machine learning (ML). The SDGs are arguably the most urgent challenges the world faces today, and it is important that the ML community contribute to solving these global issues. As progress towards SDGs is often hindered by a lack of ground survey data especially in low-income countries, ML algorithms designed for monitoring SDGs are important for leveraging non-traditional data sources that are cheap, globally available, and frequently-updated to fill in data gaps. ML-based estimates provide policymakers from governments and aid organizations with more frequent and comprehensive insights~\cite{yeh2020using,burke2021using,jean2016combining}.

The tasks defined in \bench{} can directly translate into real-world impact. For example, during the COVID-19 pandemic, the government of Togo collaborated with researchers to use satellite imagery, phone data, and ML to map poverty \cite{blumenstock2020machine} and cropland \cite{kerner2020rapid} in order to target cash payments to the jobless. Recent work in Uganda demonstrates how ML-based poverty maps can be used to measure the effectiveness of large-scale infrastructure investments \cite{ratledge2021using}. ML-based analyses of satellite images in Kenya (using the labels described in \Cref{sec:sdg2:croptype}) were recently used  to identify soil nitrogen deficiency as the limiting factor in maize yields, thereby facilitating targeted agriculture intervention \cite{jin2019smallholder}. And as a last example, the development of a new attention-based neural network architecture enabled the delineation of 1.7 million fields in Australia from satellite imagery \cite{waldner2021detect}. These field boundaries have been productized and facilitate the adoption of digital agriculture, which can improve yields while minimizing environmental pollution \cite{csiro2021epaddocks}.

Although ML approaches have demonstrated value on a variety of tasks related to SDGs~\cite{yeh2020using,burke2021using,lee2021predicting,jean2019tile2vec,jean2016combining,wang2018transfer,wang2020mapping}, the ``big data approach'' has its limits. ML models may not completely replace ground surveys. Imperfect predictions from ML models may introduce biases that propagate through downstream policy decisions, leading to negative societal impacts. The use of survey data, high resolution remote sensing images, and street-level images may also raise privacy concerns, despite efforts to protect individual privacy. We refer the reader to \Cref{app:sec:ethics} for a detailed treatment of ethical concerns in \bench, including mitigation strategies we implemented. Despite these limitations, ML applications have the greatest potential for positive impact in low-income countries, where gaps in monitoring SDGs are widest due to the constant lack of survey data.

While \bench{} is the largest SDG-focused ML dataset and benchmark to date, it is by no means complete. Field surveys are extremely costly, and labeling images for model training requires significant manual effort by experts, limiting the amount of data released in \bench{} to quantities smaller than those of many canonical ML datasets (\emph{e.g.}, ImageNet). In addition, many SDGs and indicators are not included in the current version. Such SDG indicators can be placed into 3 categories. First, several tasks can be included in future versions of \bench{} by drawing on existing data. For example, measures of gender equality (SDG 5) and access to affordable and clean energy (SDG 7) already exist in the surveys used to create labels for \bench{} tasks but will require additional processing before releasing. Recent works have also pioneered deep learning methods for identifying illegal fishing from satellite images~\cite{park2020illuminating} (SDG 14) and monitoring biodiversity from camera traps~\cite{beery2020iwildcam} (SDG 15). \Cref{tab:compare_benchmarks} includes a few relevant datasets from this first category. Second, some SDG indicators require additional research to discover non-traditional data modalities that can be used to monitor them. Finally, not all SDGs are measurable using ML or need improved measurement capabilities from ML models. For example, international cooperation (SDG 17) is perhaps best measured by domestic and international policies and agreements.

For the ML community, \bench{} also provides opportunities to test state-of-the-art ML models on real-world data and develop novel algorithms. For example, the tasks based on DHS household survey data share the same inputs and thus facilitate multi-task training. In particular, we encourage researchers to take advantage of the known strong associations between asset wealth, child mortality, women's education, and sanitation labels \cite{fink2011effect,graetz2018mapping}. The combination of satellite and street-level imagery for these tasks also enables multi-modal representation learning. On the other hand, the land cover classification and cropland mapping tasks provide new real-world datasets for evaluating and developing self-supervised, weakly supervised, unsupervised, and meta-learning algorithms. We welcome exploration of methods beyond our provided baseline models.

Ultimately, we hope \bench{} will lower the barrier to entry for the ML community to contribute toward monitoring SDGs and highlight challenges for ML researchers to address. In the long run, we plan to continue expanding datasets and benchmarks as new data sources become available. We believe that standardized datasets and benchmarks like those in \bench{} are imperative to both novel method development and real-world impact.

\section*{Acknowledgments}
The authors would like to thank everyone from the Stanford Sustainability and AI Lab for constructive feedback and discussion; the Mapillary team for technical support on the dataset; Rose Rustowicz for helping compile the crop type mapping dataset in Ghana and South Sudan; Anna X. Wang and Jiaxuan You for their help in making the crop yield dataset; and Han Lin Aung and Burak Uzkent for permission to release the field delineation dataset.

This work was supported by NSF awards (\#1651565, \#1522054), the Stanford Institute for Human-Centered AI (HAI), the Stanford King Center, the United States Agency for International Development (USAID), a Sloan Research Fellowship, and the Global Innovation Fund.

\def\bibfont{\small}
\bibliography{sources}

\newpage
\appendix
\renewcommand{\thefigure}{A\arabic{figure}}
\renewcommand{\thetable}{A\arabic{table}}
\setcounter{figure}{0}
\setcounter{table}{0}

\section*{Appendix}
\label{sec:app}

\section{Dataset Licenses}
\label{app:sec:licenses}
The Landsat, DMSP, NAIP, and VIIRS satellite images provided in SustainBench are in the public domain. PlanetScope imagery and Mapillary street-level imagery are provided under the CC BY-SA 4.0 license. Sentinel-2 imagery is provided under the Open Access compliant Creative Commons CC BY-SA 3.0 IGO license. Sentinel-1 imagery provides free access to imagery, including reproduction and distribution \footnote{\url{https://scihub.copernicus.eu/twiki/pub/SciHubWebPortal/TermsConditions/Sentinel_Data_Terms_and_Conditions.pdf}}. Likewise, MODIS imagery is free to reuse and redistribute \footnote{\url{https://lpdaac.usgs.gov/data/data-citation-and-policies/}}.

Our inclusion of labels derived from DHS survey data is within the DHS program Terms of Use\footnote{\url{https://dhsprogram.com/data/terms-of-use.cfm}} as the labels are aggregated to the cluster level and do not include any of the original ``micro-level'' data, and no individuals are identified.

Our inclusion of labels derived from LSMS survey data is within the LSMS access policy, as we do not redistribute any of the raw data files.

The Argentina crop yield labels are provided under the CC BY 2.5 AR license. United States crop yield labels are also free to access and reproduce \footnote{\url{https://www.nass.usda.gov/Data_and_Statistics/Citation_Request/index.php}}.

The brick kiln binary classification labels were manually hand-labeled by ourselves and our collaborators and therefore do not have any licensing restrictions.

\bench{} itself is released under a CC BY-SA 4.0 license, which is compatible with all of the licenses for the datasets included.

\section{Dataset Storage and Maintenance Plans}
\label{app:dataset_storage_plans}

Our datasets are stored on Google Drive at the following link:
\url{https://drive.google.com/drive/folders/1jyjK5sKGYegfHDjuVBSxCoj49TD830wL?usp=sharing}. Due to the large size of our dataset, we were unable to find any existing research data repository (\emph{e.g.}, Zenodo, Dataverse) willing to accommodate our dataset.

The GitHub repo with code used to process the datasets and run our baseline models is located at \url{https://github.com/sustainlab-group/sustainbench/}.

The dataset will be maintained by the Stanford Sustainability and AI lab.

\newpage
\section{The 17 Sustainable Development Goals (SDGs)}

\begin{table}
\centering
\caption{The full list of 17 UN Sustainable Development Goals (SDGs), along with the number of targets and indicators divided by tier.}
\label{app:tab:sdgs_full}
\begin{adjustbox}{max width=\linewidth}
\begin{tabular}{c >{\raggedright\arraybackslash}p{0.18\textwidth} >{\raggedright\arraybackslash}p{0.40\textwidth} c c c c}
    \toprule
    \textbf{SDG} & \textbf{Name} & \textbf{Description} & \textbf{\# of} & \multicolumn{3}{c}{\textbf{\# of Indicators}} \\
    \cline{5-7}
    \textbf{\#} & & & \textbf{Targets} & Tier I & Tier II & Tier I/II \\
    \midrule
    1 & No Poverty & End poverty in all its forms everywhere & 7 & 5 & 8 & 0 \\
    2 & Zero Hunger & End hunger, achieve food security and improved nutrition and promote sustainable agriculture & 8 & 10 & 4 & 0 \\
    3 & Good Health and Well-Being & Ensure healthy lives and promote well-being for all at all ages & 13 & 25 & 3 & 0 \\
    4 & Quality Education & Ensure inclusive and equitable quality education and promote lifelong learning opportunities for all & 10 & 5 & 6 & 1 \\
    5 & Gender Equality & Achieve gender equality and empower all women and girls & 9 & 4 & 10 & 0 \\
    6 & Clean Water and Sanitation & Ensure availability and sustainable management of water and sanitation for all & 8 & 7 & 4 & 0 \\
    7 & Affordable and Clean Energy & Ensure access to affordable, reliable, sustainable and modern energy for all & 5 & 6 & 0 & 0 \\
    8 & Decent Work and Economic Growth & Promote sustained, inclusive and sustainable economic growth, full and productive employment and decent work for all & 12 & 8 & 8 & 0 \\
    9 & Industry, Innovation and Infrastructure & Build resilient infrastructure, promote inclusive and sustainable industrialization and foster innovation & 8 & 10 & 2 & 0 \\
    10 & Reduced Inequalities & Reduce inequality within and among countries & 10 & 8 & 6 & 0 \\
    11 & Sustainable Cities and Communities & Make cities and human settlements inclusive, safe, resilient and sustainable & 10 & 4 & 10 & 0 \\
    12 & Responsible Consumption and Production & Ensure sustainable consumption and production patterns & 11 & 5 & 8 & 0 \\
    13 & Climate Action & Take urgent action to combat climate change and its impacts & 5 & 2 & 6 & 0 \\
    14 & Life below Water & Conserve and sustainably use the oceans, seas and marine resources for sustainable development & 10 & 5 & 5 & 0 \\
    15 & Life on Land & Protect, restore and promote sustainable use of terrestrial ecosystems, sustainably manage forests, combat desertification, and halt and reverse land degradation and halt biodiversity loss & 12 & 11 & 2 & 1 \\
    16 & Peace, Justice and Strong Institutions &Promote peaceful and inclusive societies for sustainable development, provide access to justice for all and build effective, accountable and inclusive institutions at all levels & 12 & 6 & 17 & 1 \\
    17 & Partnerships for the Goals & Strengthen the means of implementation and revitalize the global partnership for sustainable development & 19 & 15 & 8 & 1 \\
    \midrule
    \textbf{Total} & & & 169 & 136 & 107 & 4 \\
    \bottomrule
\end{tabular}
\end{adjustbox}
\end{table}

Today, six years after the unveiling of the SDGs, many gaps still exist in monitoring progress. Official tracking of data availability is conducted by the UN Statistical Commission, which classifies each
indicator into one of three tiers: indicator is well-defined and data are regularly produced by at least 50\% of countries (Tier I), indicator is well-defined but data are not regularly produced by countries (Tier II), and the indicator is currently not well-defined (Tier III). As of the latest report from March 2021, 136 indicators have regular data from at least 50\% of countries, 107 indicators have sporadic data, and 4 indicators are a mix depending on the data of interest (\Cref{app:tab:sdgs_full}) \cite{un2021tier}. For example, for monitoring global poverty (SDG 1), the proportion of a country's population living below the international poverty line (Indicator 1.1.1) is reported annually for all countries, but the economic loss attributed to natural and man-made disasters (Indicator 1.5.2) is only sparsely documented. We provide descriptions of the 17 Sustainable Development Goals (SDGs) in \Cref{app:tab:sdgs_full}.

\section{Dataset Details}
\label{app:sec:dataset}

\begin{figure}
\centering
\begin{subfigure}[b]{0.45\textwidth}
  \centering
  \includegraphics[trim={5cm 5cm 5cm 5cm},clip,width=\textwidth]{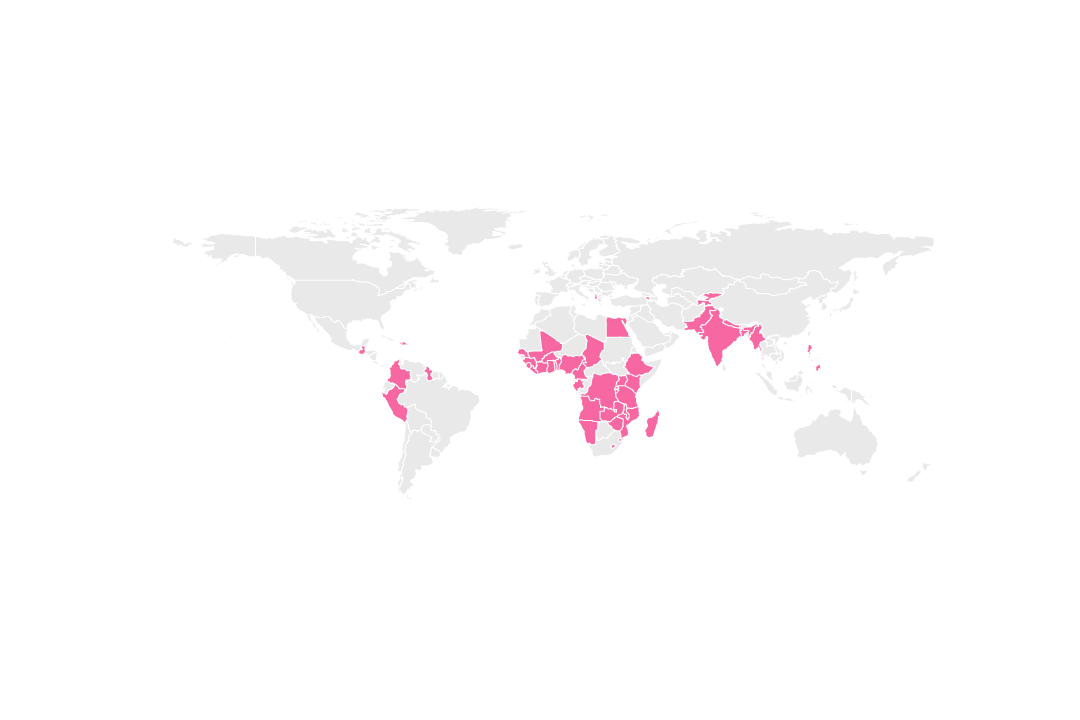}
  \caption{SDG 1: No Poverty \\ (48 countries)}
\end{subfigure}
\begin{subfigure}[b]{0.45\textwidth}
  \centering
  \includegraphics[trim={5cm 5cm 5cm 5cm},clip,width=\textwidth]{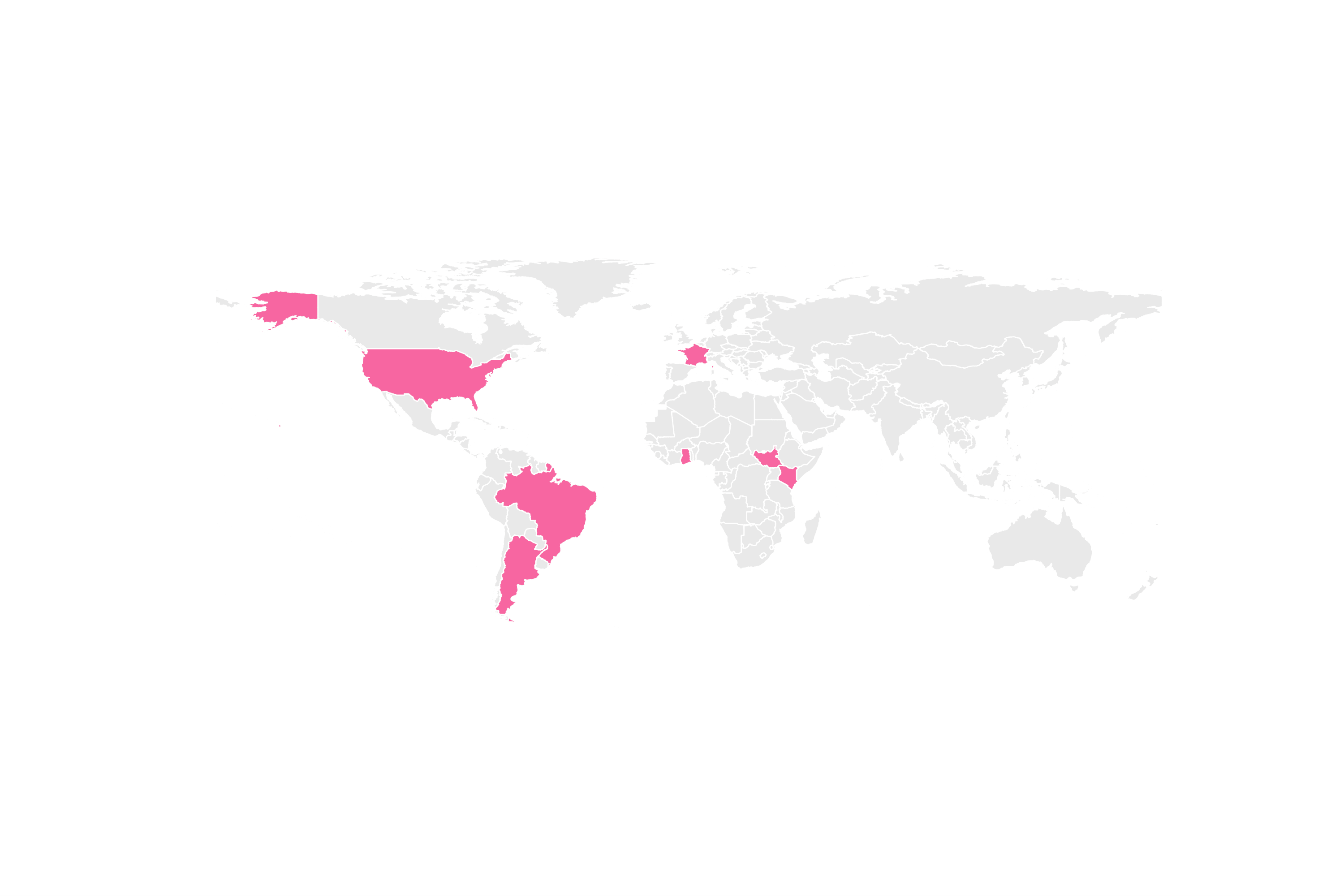}
  \caption{SDG 2: No Hunger \\ (7 countries)}
\end{subfigure}
\newline
\begin{subfigure}[b]{0.45\textwidth}
  \centering
  \captionsetup{justification=centering}
  \includegraphics[trim={5cm 5cm 5cm 5cm},clip,width=\textwidth]{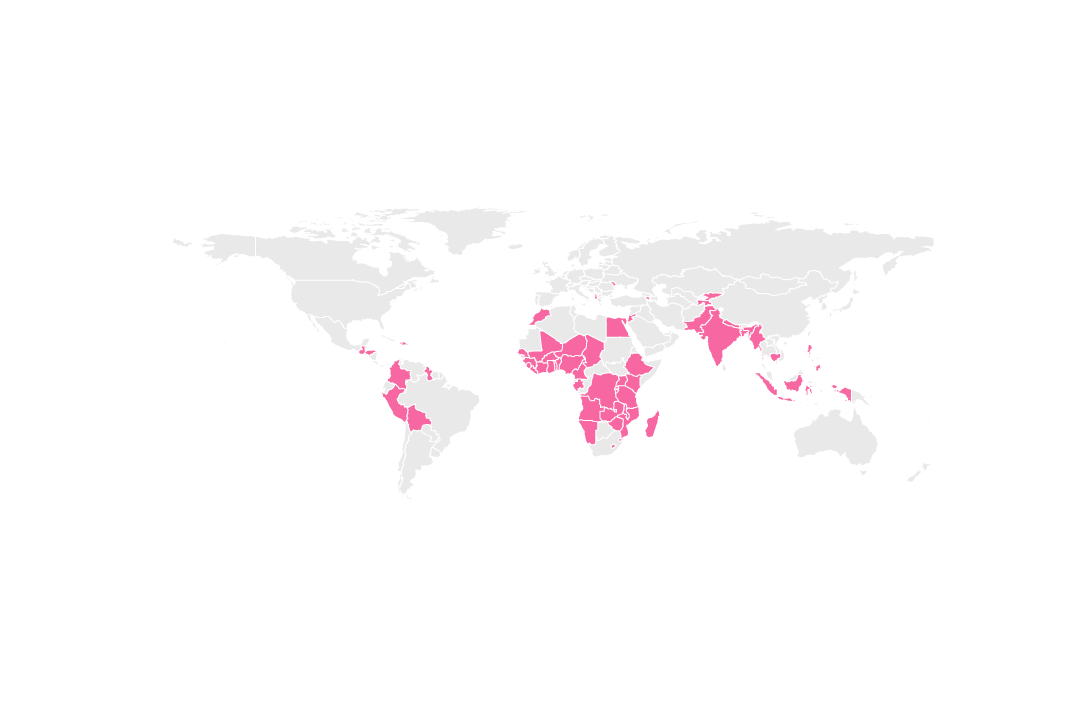}
  \caption{SDG 3: Good Health and Well-Being \\ (56 countries)}
\end{subfigure}
\begin{subfigure}[b]{0.45\textwidth}
  \centering
  \includegraphics[trim={5cm 5cm 5cm 5cm},clip,width=\textwidth]{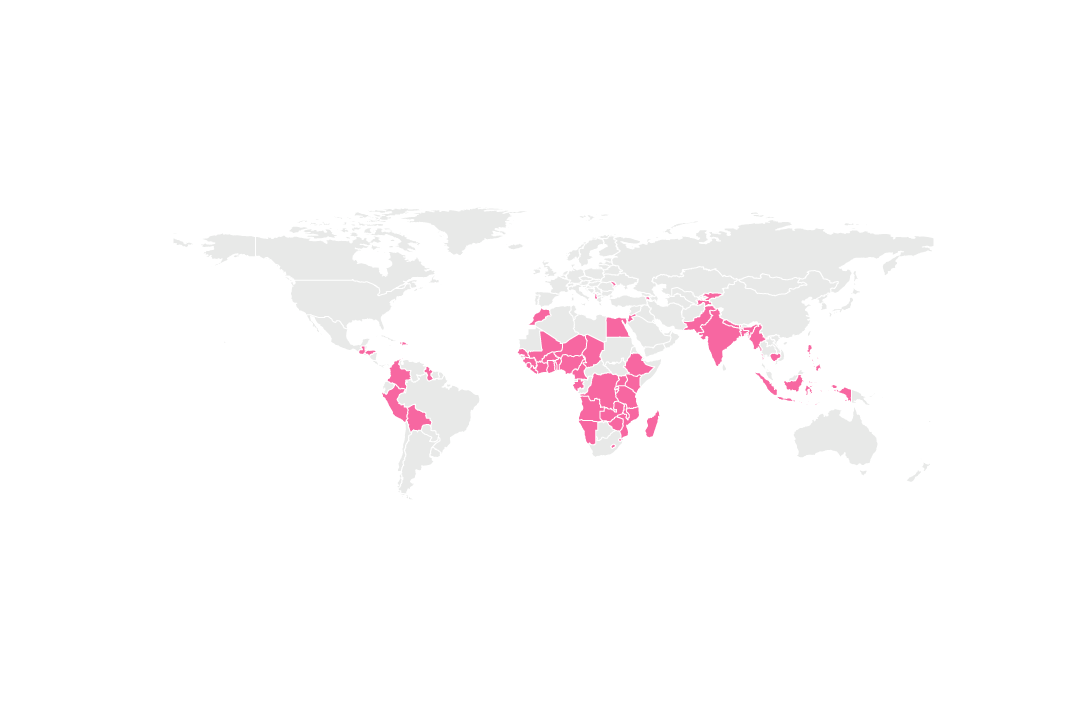}
  \caption{SDG 4: Quality Education \\ (56 countries)}
\end{subfigure}
\newline
\begin{subfigure}[b]{0.45\textwidth}
  \centering
  \captionsetup{justification=centering}
  \includegraphics[trim={5cm 5cm 5cm 5cm},clip,width=\textwidth]{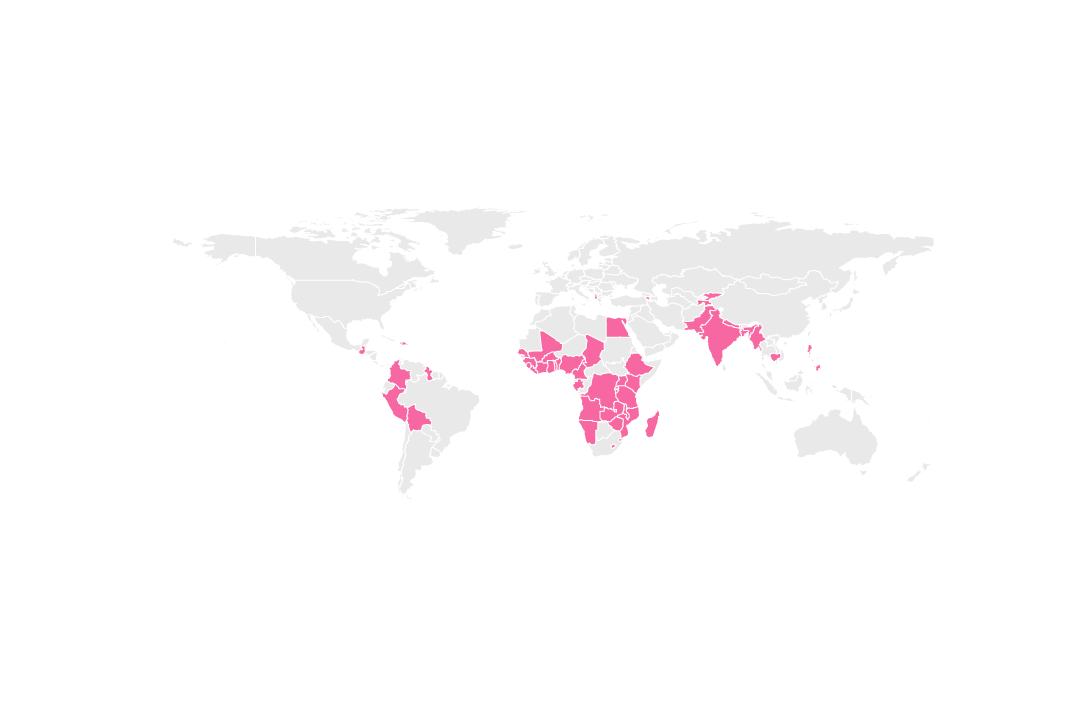}
  \caption{SDG 6: Clean Water and Sanitation \\ (50 countries)}
\end{subfigure}
\begin{subfigure}[b]{0.45\textwidth}
  \centering
  \includegraphics[trim={5cm 5cm 5cm 5cm},clip,width=\textwidth]{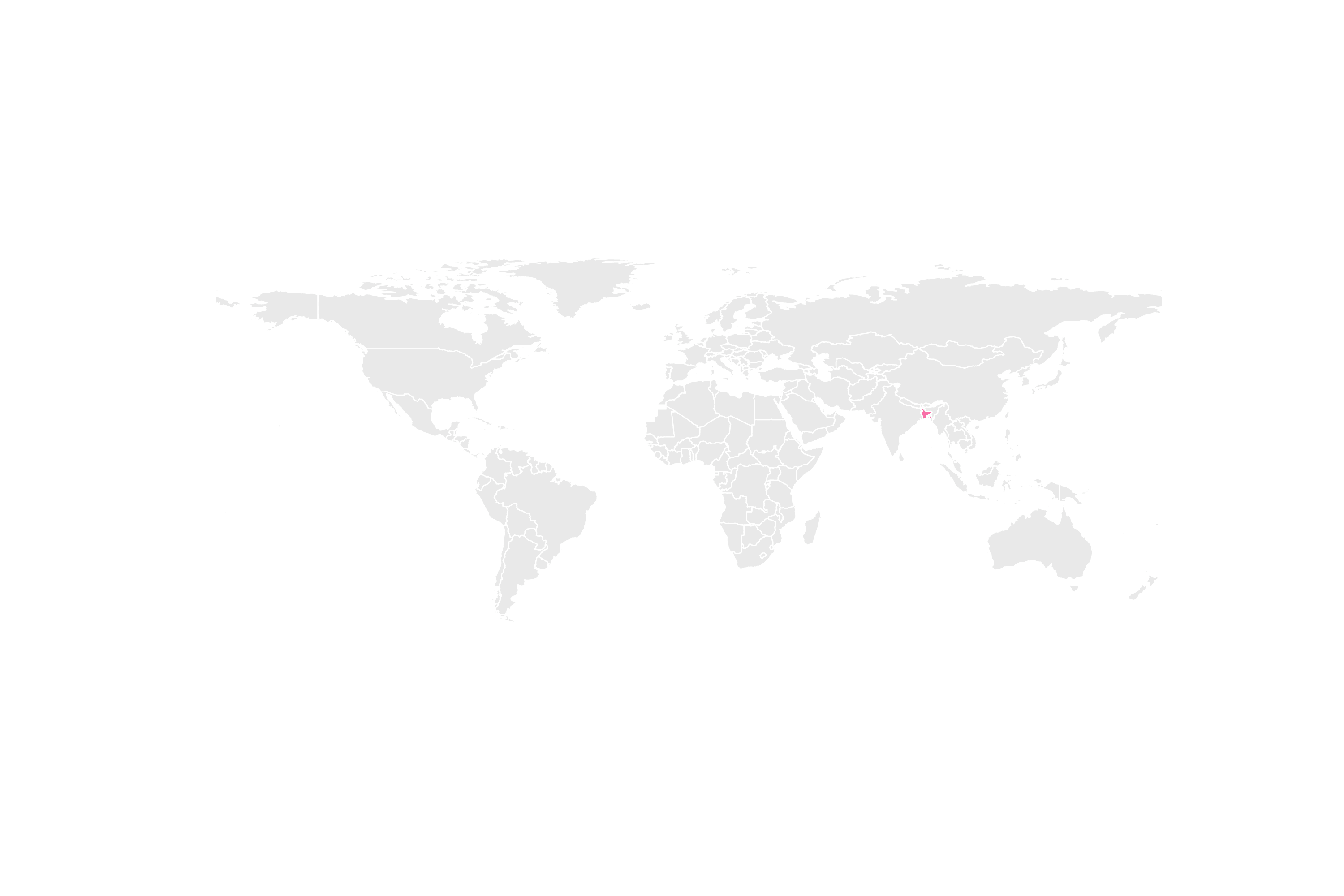}
  \caption{SDG 13: Climate Action \\ (1 country)}
\end{subfigure}
\newline
\begin{subfigure}[b]{0.45\textwidth}
  \centering
  \includegraphics[trim={5cm 5cm 5cm 5cm},clip,width=\textwidth]{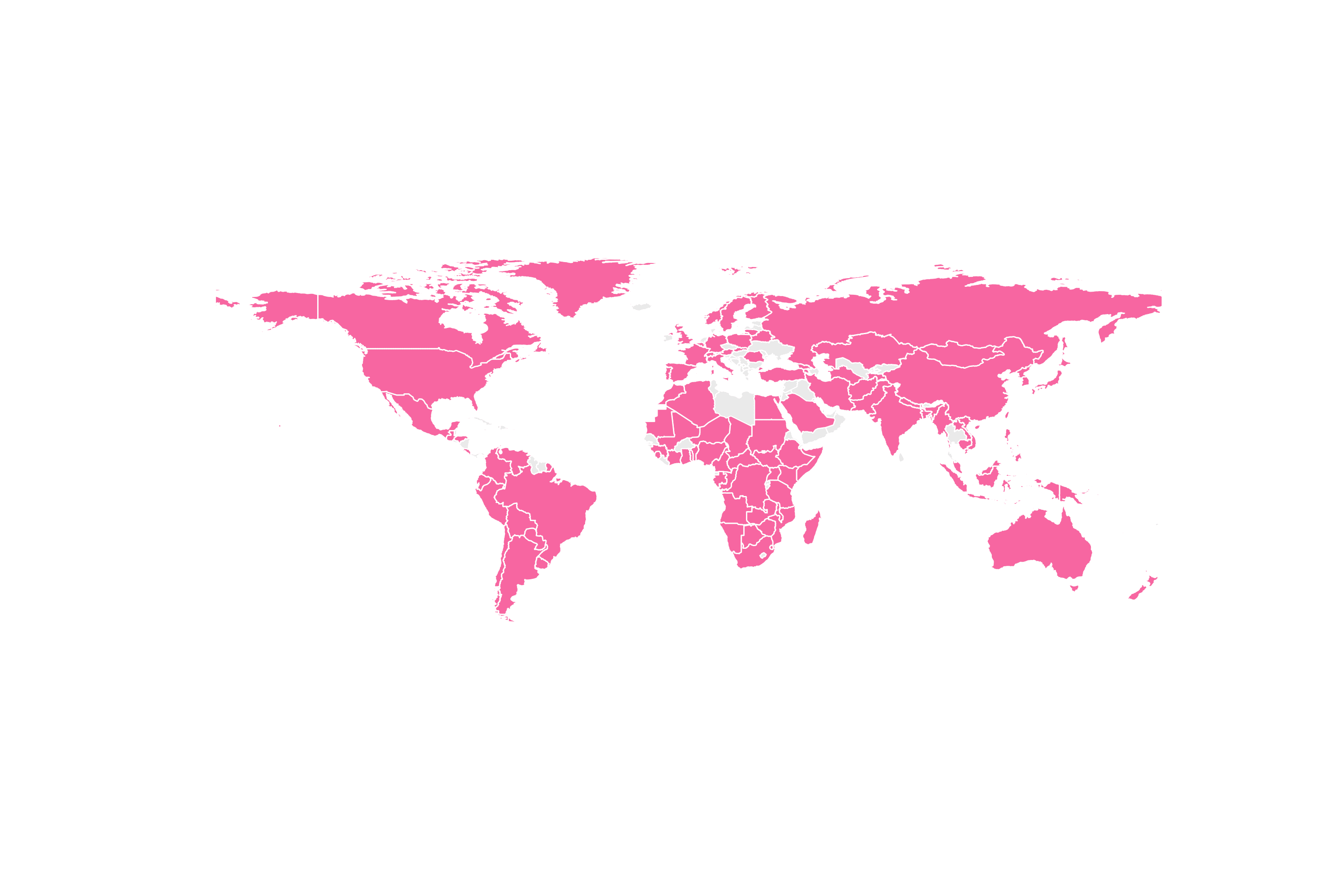}
  \caption{SDG 15: Life on Land \\ (105 countries)}
\end{subfigure}
\caption{Maps of geographic \bench{} coverage per SDG.}
\label{app:fig:sdgmaps}
\end{figure}

\subsection{DHS-based datasets}
\label{app:dhs}

In this section, we detail the process of constructing the poverty, health, education, and water and sanitation labels from DHS surveys. We also give more information about the input imagery that we provide as part of \bench.

\paragraph{Labels from DHS survey data}

We constructed several indices using survey data from the Demographic and Health Surveys (DHS) program, which is funded by the US Agency for International Development (USAID) and has conducted nationally representative household-level surveys in over 90 countries. For \bench, we combined survey data covering 56 countries from 179 unique surveys with questions on women's education, women's BMI, under 5 mortality, household asset ownership, water quality, and sanitation (toilet) quality. We chose surveys between 1996 (the first year that nightlights imagery is available) and 2019 (the latest year with available DHS surveys)\footnote{Even though a DHS survey may have been conducted over several years, we refer to the ``year'' of a DHS survey as the year reported for that survey in the DHS Data API: \url{https://api.dhsprogram.com/}} for which geographic data was available. The full list of surveys is shown in \Cref{tab:dhs_surveys}.

\begin{itemize}
    \item \textbf{Asset Wealth Index}

    While the SDG indicators define poverty lines expressed in average expenditure (a.k.a. consumption) per day, survey data is much more widely available for household asset wealth than expenditure. Furthermore, asset wealth is considered a less noisy measure of households' long-run economic well-being~\cite{sahn2003exploring,filmer2012assessing} and is actively used for targeting social programs~\cite{filmer2012assessing,alkire2015multidimensional}. To summarize household-level survey data into a scalar asset wealth index, standard approaches perform principal components analysis (PCA) of survey responses and project them onto the first principal component~\cite{filmer2001estimating,sahn2003exploring}. The household-level asset wealth index is commonly averaged to create a cluster-level index, where a ``cluster'' roughly corresponds to a village or local community.

    The asset wealth index is built using household asset ownership and infrastructure information as done in prior works \cite{yeh2020using}. We include the number of rooms used for sleeping in a home (capped at 25); binary indicators for whether the household has electricity and owns a radio, TV, refrigerator, motorcycle, car, or phone (or cellphone); and the quality of floors, water source, and toilet. As ``floor type'', ``water source type'', and ``toilet type'' are reported from DHS as descriptive categorical variables (\emph{e.g.}, ``piped water''/``flush to pit latrine''), we convert the descriptions to a numeric scale, a standard technique for processing survey data \cite{deshpande2020mapping}. We use a 1-5 scale where lower numbers indicate the water source is less developed (\emph{e.g.}, straight from a lake) while higher numbers indicate higher levels of technology/development (\emph{e.g.}, piped water); we use a similar 1-5 scale for toilet type and floor type. To calculate the index, we use the first principal component of all the variables mentioned above at a household level, and report the mean at a cluster level. The asset wealth index calculation includes 2,081,808 households total from 87,119 clusters in 48 countries, with a median of 22 households per cluster. Many surveys are dropped because they do not include one of the 12 variables we use to construct the index. The final number of clusters with asset wealth labels in \bench{} is only 86,936, as several clusters did not have corresponding satellite imagery inputs. Note that households from these clusters with missing imagery still contributed to the PCA computation, since these clusters were excluded from \bench{} only \emph{after} the PCA-based index had already been constructed.

    \item \textbf{Education}

    The women's education metric is created by taking the cluster level mean of ``education in single years''. Following \cite{graetz2018mapping}, we capped the years of education at 18, a common threshold in many surveys which helps avoid outliers. The women's education metric includes data from 2,910,286 women in 56 countries, with a median of 24 women per cluster.

    \item \textbf{Health}

    To create the women's BMI metric, we first exclude all pregnant women, as the BMI is not adjusted for them. Using the sample of women BMI is appropriate for, we take the cluster level mean of reported BMI/100 (as ``decimal points are not included'' in the DHS data). The women's BMI metric includes data from 1,781,403 women in 53 countries, with a median of 18 women per cluster.

    To create the child mortality metric, we used woman level birth records. For each woman, the DHS reports up to 20 births as well as pregnancy, postnatal care, and health outcomes for each birth. Treating each child (rather than woman) as a record, we keep only the children who were age 5 or younger at the time of survey or who had died (age 5 or younger) no earlier than the year prior to the survey. After identifying the qualifying children, we calculate the number of deaths per 1,000 children by cluster. The child mortality metric includes 1,936,904 children in 56 countries, with a median of 15 children per cluster.

    \item \textbf{Water and Sanitation Indices}

    The water and sanitation indices are calculated as the cluster-level mean of our ranking of water quality and toilet type, respectively. The water index calculation includes 2,105,026 households over 49 countries, with a median of 22 households per cluster. The sanitation index calculation includes 2,143,329 households over 49 countries, with a median of 22 households per cluster.
\end{itemize}

For all indices, we excluded the calculated index for a cluster if fewer than 5 observations are used to create it. For the asset wealth, sanitation, and water indices an observation unit is a household; for the women's education, BMI and under 5 mortality measures the observation unit is an individual. We also excluded several hundred clusters for which satellite imagery could not be obtained.

For all of the tasks based on DHS survey data, we use a uniform train/validation/test dataset split by country. Delineating by country ensures that there is no overlap between any of the splits---\emph{i.e.}, a model trained on our train split will not have ``seen'' any part of any image from the test split. The splits are listed in \Cref{app:tab:dhs_splits}.

\begin{table}
\centering
\caption{Splits for DHS survey-based tasks. See \Cref{tab:dhs_surveys} for the mapping between DHS country code and the full country name.}
\label{app:tab:dhs_splits}
\begin{adjustbox}{max width=\linewidth}
\begin{tabular}{>{\raggedright\arraybackslash}p{0.2\linewidth} >{\raggedright\arraybackslash}p{0.235\linewidth} >{\raggedright\arraybackslash}p{0.235\linewidth} >{\raggedright\arraybackslash}p{0.235\linewidth}}
\toprule
    & Train & Validation & Test \\
\midrule
    DHS Country Codes
        & 30 countries: \hspace{3em} \texttt{AL, BD, CD, CM, GH, GU, HN, IA, ID, JO, KE, KM, LB, LS, MA, MB, MD, MM, MW, MZ, NG, NI, PE, PH, SN, TG, TJ, UG, ZM, ZW}
        & 13 countries: \hspace{3em} \texttt{BF, BJ, BO, CO, DR, GA, GN, GY, HT, NM, SL, TD, TZ}
        & 13 countries: \hspace{3em} \texttt{AM, AO, BU, CI, EG, ET, KH, KY, ML, NP, PK, RW, SZ} \\
    asset wealth index   & 59,617 examples (69\%) & 16,776 examples (19\%) & 10,543 examples (12\%) \\
    child mortality rate & 69,052 (65\%) & 17,062 (16\%) & 19,468 (18\%) \\
    women BMI            & 61,950 (65\%) & 15,675 (17\%) & 17,241 (18\%) \\
    women education      & 75,818 (65\%) & 20,589 (18\%) & 20,655 (18\%) \\
    water index          & 59,620 (68\%) & 17,773 (20\%) & 10,545 (12\%) \\
    sanitation index     & 60,184 (67\%) & 16,776 (19\%) & 12,311 (14\%) \\
\bottomrule
\end{tabular}
\end{adjustbox}
\end{table}

\begingroup
\begin{table}[htbp]
\centering
\caption{179 DHS surveys from 56 countries spanning 1996-2019 were used to create labels.}
\label{tab:dhs_surveys}
\renewcommand{\arraystretch}{0.85} %
\begin{tabular}{>{\footnotesize}l>{\scriptsize\raggedright\arraybackslash}p{0.62\linewidth}}
    \toprule
\small{DHS Code - Country}     & \small{Survey IDs (\texttt{SurveyId} field from the DHS Data API)} \\
\midrule
\texttt{AL} - Albania                   & \texttt{AL2008DHS}, \texttt{AL2017DHS} \\
\texttt{AM} - Armenia                   & \texttt{AM2010DHS}, \texttt{AM2016DHS} \\
\texttt{AO} - Angola                    & \texttt{AO2006MIS}, \texttt{AO2011MIS}, \texttt{AO2015DHS} \\
\texttt{BD} - Bangladesh                & \texttt{BD2000DHS}, \texttt{BD2004DHS}, \texttt{BD2007DHS}, \texttt{BD2011DHS}, \texttt{BD2014DHS}, \texttt{BD2017DHS} \\
\texttt{BF} - Burkina Faso              & \texttt{BF1999DHS}, \texttt{BF2003DHS}, \texttt{BF2010DHS}, \texttt{BF2014MIS}, \texttt{BF2017MIS} \\
\texttt{BJ} - Benin                     & \texttt{BJ1996DHS}, \texttt{BJ2001DHS}, \texttt{BJ2012DHS}, \texttt{BJ2017DHS} \\
\texttt{BO} - Bolivia                   & \texttt{BO2008DHS} \\
\texttt{BU} - Burundi                   & \texttt{BU2010DHS}, \texttt{BU2012MIS}, \texttt{BU2016DHS} \\
\texttt{CD} - Congo Democratic Republic & \texttt{CD2007DHS}, \texttt{CD2013DHS} \\
\texttt{CI} - Cote d'Ivoire             & \texttt{CI1998DHS}, \texttt{CI2012DHS} \\
\texttt{CM} - Cameroon                  & \texttt{CM2004DHS}, \texttt{CM2011DHS}, \texttt{CM2018DHS} \\
\texttt{CO} - Colombia                  & \texttt{CO2010DHS} \\
\texttt{DR} - Dominican Republic        & \texttt{DR2007DHS}, \texttt{DR2013DHS} \\
\texttt{EG} - Egypt                     & \texttt{EG2000DHS}, \texttt{EG2003DHS}, \texttt{EG2005DHS}, \texttt{EG2008DHS}, \texttt{EG2014DHS} \\
\texttt{ET} - Ethiopia                  & \texttt{ET2000DHS}, \texttt{ET2005DHS}, \texttt{ET2011DHS}, \texttt{ET2016DHS}, \texttt{ET2019DHS} \\
\texttt{GA} - Gabon                     & \texttt{GA2012DHS} \\
\texttt{GH} - Ghana                     & \texttt{GH1998DHS}, \texttt{GH2003DHS}, \texttt{GH2008DHS}, \texttt{GH2014DHS}, \texttt{GH2016MIS}, \texttt{GH2019MIS} \\
\texttt{GN} - Guinea                    & \texttt{GN1999DHS}, \texttt{GN2005DHS}, \texttt{GN2012DHS}, \texttt{GN2018DHS} \\
\texttt{GU} - Guatemala                 & \texttt{GU2015DHS} \\
\texttt{GY} - Guyana                    & \texttt{GY2009DHS} \\
\texttt{HN} - Honduras                  & \texttt{HN2011DHS} \\
\texttt{HT} - Haiti                     & \texttt{HT2000DHS}, \texttt{HT2006DHS}, \texttt{HT2012DHS}, \texttt{HT2016DHS} \\
\texttt{IA} - India                     & \texttt{IA2015DHS} \\
\texttt{ID} - Indonesia                 & \texttt{ID2003DHS} \\
\texttt{JO} - Jordan                    & \texttt{JO2002DHS}, \texttt{JO2007DHS}, \texttt{JO2012DHS}, \texttt{JO2017DHS} \\
\texttt{KE} - Kenya                     & \texttt{KE2008DHS}, \texttt{KE2014DHS}, \texttt{KE2015MIS} \\
\texttt{KH} - Cambodia                  & \texttt{KH2000DHS}, \texttt{KH2005DHS}, \texttt{KH2010DHS}, \texttt{KH2014DHS} \\
\texttt{KM} - Comoros                   & \texttt{KM2012DHS} \\
\texttt{KY} - Kyrgyz Republic           & \texttt{KY2012DHS} \\
\texttt{LB} - Liberia                   & \texttt{LB2007DHS}, \texttt{LB2009MIS}, \texttt{LB2011MIS}, \texttt{LB2013DHS}, \texttt{LB2016MIS}, \texttt{LB2019DHS} \\
\texttt{LS} - Lesotho                   & \texttt{LS2004DHS}, \texttt{LS2009DHS}, \texttt{LS2014DHS} \\
\texttt{MA} - Morocco                   & \texttt{MA2003DHS} \\
\texttt{MB} - Moldova                   & \texttt{MB2005DHS} \\
\texttt{MD} - Madagascar                & \texttt{MD1997DHS}, \texttt{MD2008DHS}, \texttt{MD2011MIS}, \texttt{MD2013MIS}, \texttt{MD2016MIS} \\
\texttt{ML} - Mali                      & \texttt{ML1996DHS}, \texttt{ML2001DHS}, \texttt{ML2006DHS}, \texttt{ML2012DHS}, \texttt{ML2015MIS}, \texttt{ML2018DHS} \\
\texttt{MM} - Myanmar                   & \texttt{MM2016DHS} \\
\texttt{MW} - Malawi                    & \texttt{MW2000DHS}, \texttt{MW2004DHS}, \texttt{MW2010DHS}, \texttt{MW2012MIS}, \texttt{MW2014MIS}, \texttt{MW2015DHS}, \texttt{MW2017MIS} \\
\texttt{MZ} - Mozambique                & \texttt{MZ2009AIS}, \texttt{MZ2011DHS}, \texttt{MZ2015AIS}, \texttt{MZ2018MIS} \\
\texttt{NG} - Nigeria                   & \texttt{NG2003DHS}, \texttt{NG2008DHS}, \texttt{NG2010MIS}, \texttt{NG2013DHS}, \texttt{NG2015MIS}, \texttt{NG2018DHS} \\
\texttt{NI} - Niger                     & \texttt{NI1998DHS} \\
\texttt{NM} - Namibia                   & \texttt{NM2000DHS}, \texttt{NM2006DHS}, \texttt{NM2013DHS} \\
\texttt{NP} - Nepal                     & \texttt{NP2001DHS}, \texttt{NP2006DHS}, \texttt{NP2011DHS}, \texttt{NP2016DHS} \\
\texttt{PE} - Peru                      & \texttt{PE2000DHS}, \texttt{PE2004DHS}, \texttt{PE2007DHS}, \texttt{PE2009DHS} \\
\texttt{PH} - Philippines               & \texttt{PH2003DHS}, \texttt{PH2008DHS}, \texttt{PH2017DHS} \\
\texttt{PK} - Pakistan                  & \texttt{PK2006DHS}, \texttt{PK2017DHS} \\
\texttt{RW} - Rwanda                    & \texttt{RW2005DHS}, \texttt{RW2008DHS}, \texttt{RW2010DHS}, \texttt{RW2015DHS} \\
\texttt{SL} - Sierra Leone              & \texttt{SL2008DHS}, \texttt{SL2013DHS}, \texttt{SL2016MIS}, \texttt{SL2019DHS} \\
\texttt{SN} - Senegal                   & \texttt{SN1997DHS}, \texttt{SN2005DHS}, \texttt{SN2008MIS}, \texttt{SN2010DHS}, \texttt{SN2012DHS}, \texttt{SN2015DHS}, \texttt{SN2017DHS}, \texttt{SN2018DHS} \\
\texttt{SZ} - Eswatini                  & \texttt{SZ2006DHS} \\
\texttt{TD} - Chad                      & \texttt{TD2014DHS} \\
\texttt{TG} - Togo                      & \texttt{TG1998DHS}, \texttt{TG2013DHS}, \texttt{TG2017MIS} \\
\texttt{TJ} - Tajikistan                & \texttt{TJ2012DHS}, \texttt{TJ2017DHS} \\
\texttt{TZ} - Tanzania                  & \texttt{TZ1999DHS}, \texttt{TZ2007AIS}, \texttt{TZ2010DHS}, \texttt{TZ2012AIS}, \texttt{TZ2015DHS}, \texttt{TZ2017MIS} \\
\texttt{UG} - Uganda                    & \texttt{UG2000DHS}, \texttt{UG2006DHS}, \texttt{UG2009MIS}, \texttt{UG2011DHS}, \texttt{UG2014MIS}, \texttt{UG2016DHS}, \texttt{UG2018MIS} \\
\texttt{ZM} - Zambia                    & \texttt{ZM2007DHS}, \texttt{ZM2013DHS}, \texttt{ZM2018DHS} \\
\texttt{ZW} - Zimbabwe                  & \texttt{ZW1999DHS}, \texttt{ZW2005DHS}, \texttt{ZW2010DHS}, \texttt{ZW2015DHS} \\
    \bottomrule
\end{tabular}
\end{table}
\endgroup

\paragraph{Multispectral (MS) bands}
The main source of inputs for these tasks is satellite imagery, collected and processed in a similar manner as \cite{yeh2020using}. For each DHS surveyed country and year, we created 3-year median composites of daytime surface reflectance images captured by the Landsat 5, 7, and 8 satellites. Each composite takes the median of each cloud-free pixel available during a 3-year period centered on the year of the DHS survey. (Note the difference from \cite{yeh2020using}, which only chose three distinct 3-year periods for compositing.)  As described in \cite{yeh2020using}, the motivation for using 3-year composites is two-fold. First, multi-year median compositing has seen success in similar applications for gathering clear satellite imagery \cite{azzari2017landsat}, and even in 1-year composites we observed substantial influence of clouds in some regions, given imperfections in the cloud mask. Second, the outcomes that we predict (wealth, health, education, and infrastructure) tend to evolve slowly over time, and we did not want our inputs to be distorted by seasonal or short-run variation. These daytime images have a spatial resolution of 30 m/pixel with seven bands which we refer to as the multispectral (MS) bands: \texttt{RED}, \texttt{GREEN}, \texttt{BLUE}, \texttt{NIR} (Near Infrared), \texttt{SWIR1} (Shortwave Infrared 1), \texttt{SWIR2} (Shortwave Infrared 2), and \texttt{TEMP1} (Thermal).

\paragraph{Nightlights (NL)}
We also include nighttime lights (``nightlights'') imagery, using the same sources as \cite{yeh2020using}. No single satellite captured calibrated nightlights for all of 1996-2019, so we collected DMSP-OLS Radiance Calibrated Nighttime Lights \cite{hsu2015dmsp} for the years 1996-2011, and VIIRS Nighttime Day/Night Band \cite{elvidge2017viirs} for the years 2012-2019. DMSP nightlights have 30 arc-second/pixel resolution and are considered unitless, whereas VIIRS nightlights have 15 arc-second/pixel resolution and units of radiance (nW~cm$^{-2}$~sr$^{-1}$). For the DMSP calibrated nightlights, which only exists as annual composites for a few specific years, we chose the annual composite closest to the year of the DHS survey; furthermore, we use the inter-satellite calibration procedure from \cite{hsu2015dmsp} to ensure that the DMSP values are comparable across time (a procedure which \cite{yeh2020using} did not follow). For VIIRS, which provides monthly composites, we perform 3-year median compositing similar to the Landsat images, taking the median of each monthly average radiance over a 3-year period centered on the year of the DHS survey. All nightlights images are resized using nearest-neighbor upsampling to cover the same spatial area as each Landsat image.

\begin{figure}
\centering
\includegraphics[width=0.3\textwidth]{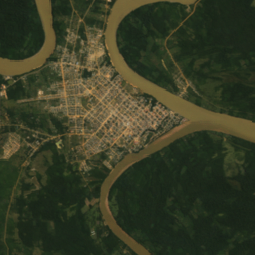}
\qquad
\includegraphics[width=0.3\textwidth]{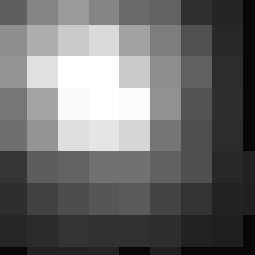}
\caption{An example of an input satellite image for the DHS survey-based datasets. This image is of cluster 969 from the 2004 DHS survey of Peru, located at latitude and longitude coordinates of (-12.597851, -69.185416). The left image shows the RGB channels from Landsat surface reflectance. The right image shows the Nightlights band from DMSP.}
\label{app:fig:landsat_nl}
\end{figure}

The MS and NL satellite imagery were processed in and exported from Google Earth Engine \cite{gorelick2017google}. For each cluster from a given DHS surveyed country-year, we provide one 255$\times$255$\times$8 image (7 MS bands, 1 NL band) centered on the cluster's geocoordinates at a scale of 30 m/pixel. See \Cref{app:fig:landsat_nl} for an example of an image in our dataset. In our released code, we provide the mean and standard deviation of each band across the entire dataset for input normalization.

The exact image collections we used on Google Earth Engine are as follows:
\begin{itemize}[nosep]
    \item USGS Landsat 5, Collection 1 Surface Reflectance Tier 1: \verb|LANDSAT/LT05/C01/T1_SR|
    \item USGS Landsat 7, Collection 1 Surface Reflectance Tier 1: \verb|LANDSAT/LE07/C01/T1_SR|
    \item USGS Landsat 8, Collection 1 Surface Reflectance Tier 1: \verb|LANDSAT/LC08/C01/T1_SR|
    \item DMSP-OLS Global Radiance-Calibrated Nighttime Lights Version 4: \\ \verb|NOAA/DMSP-OLS/CALIBRATED_LIGHTS_V4|
    \item VIIRS Nighttime Day/Night Band Composites Version 1: \\ \verb|NOAA/VIIRS/DNB/MONTHLY_V1/VCMCFG|
\end{itemize}

For future releases of \bench, we would like to update all of the Landsat imagery to the newer ``Collection 2'' products. New Collection 1 products will not be released beyond January 1, 2022, so we would not be able to use the existing Collection 1 imagery source for future DHS surveys. We would also like to update the VIIRS imagery to the official annual composites released by the Earth Observation Group. We did not provide such imagery in \bench{} because they were not available on Google Earth Engine at the time \bench{} was compiled.

\begin{figure}
\centering
\includegraphics[width=0.3\textwidth]{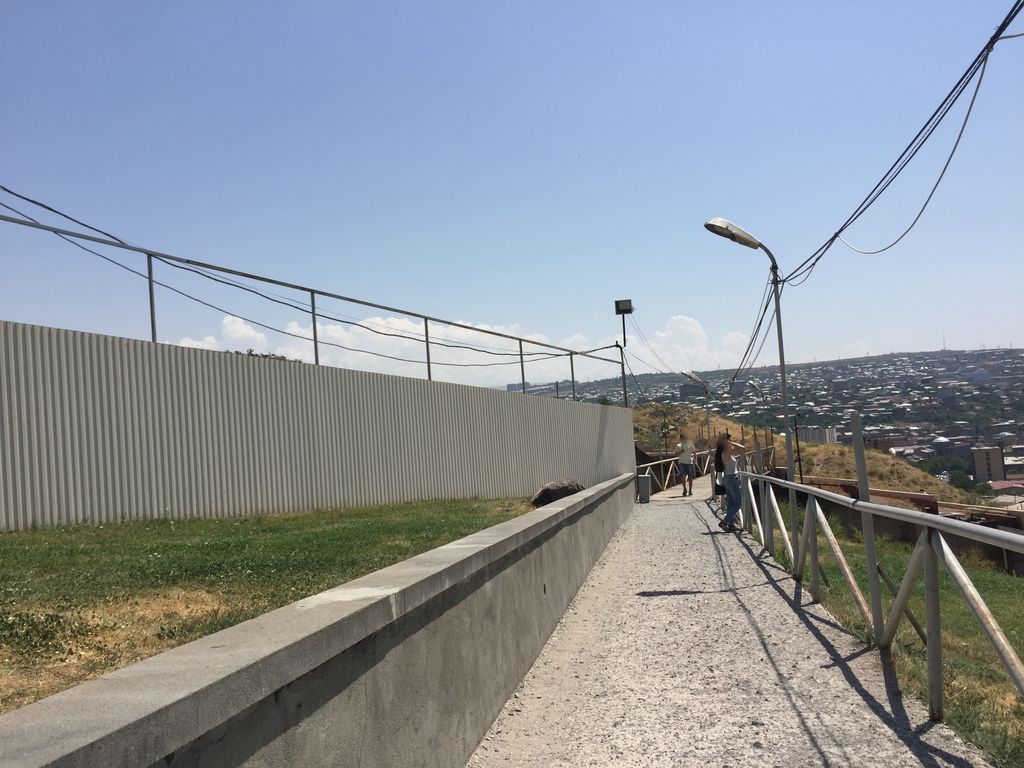}
\qquad
\includegraphics[width=0.3\textwidth]{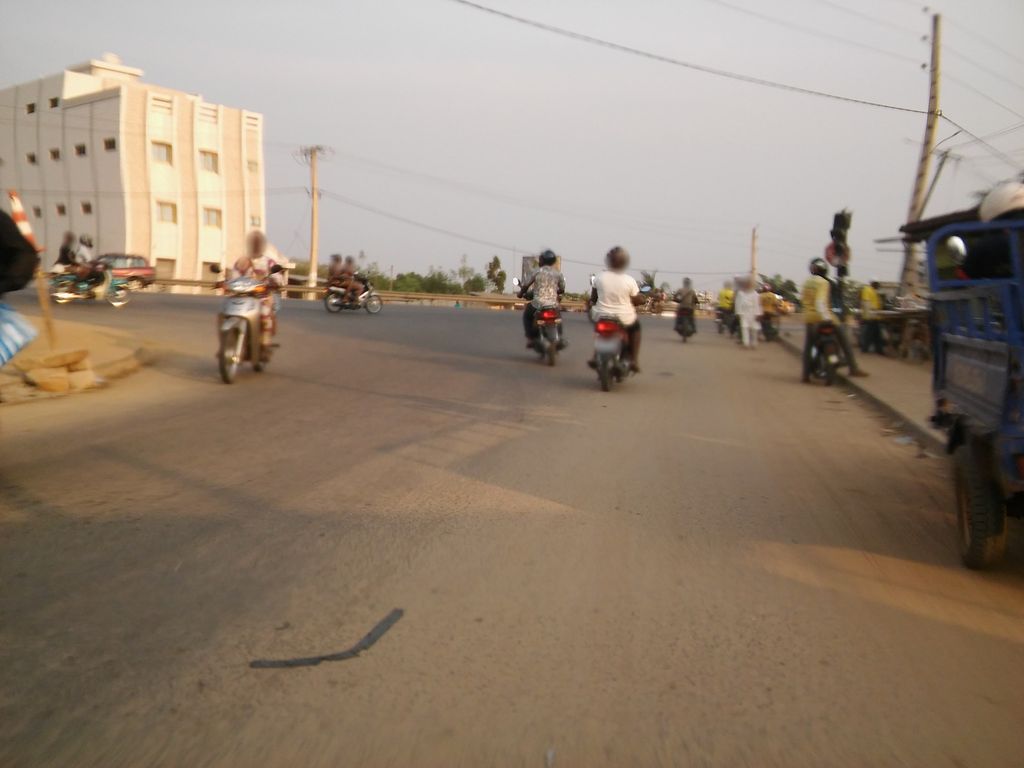}
\caption{An example of an input street-level image from Mapillary for the DHS survey-based datasets. The left image is from cluster 10 of Armenia located at (40.192860, 44.515051). The right image is from cluster 92 of Benin, located at (2.347327, 6.402679).}
\label{app:fig:mapillary}
\end{figure}

\paragraph{Mapillary Images}
Mapillary~\cite{MVD2017} provides a platform for crowd-sourced, geo-tagged street-level imagery. It provides an API to access data such as images, map features, and object detections, automatically blurring faces of human subjects and license places \cite{BlurringImagesModelNeuhold} and allowing users who upload images to manually blur if any are missed \cite{BlurringImages} for privacy. We retrieved only images that intersect with a DHS cluster. A given image must satisfy two conditions to intersect with a DHS cluster: 1) its geo-coordinates must be within 0.1 degree latitude and longitude to the cluster's geo-location, and 2) it must have been captured within 3 years before or after the year of the DHS datapoint. Each image has metadata, including a unique ID, timestamp of capture in milliseconds, year of capture, latitude, and longitude. All downloaded images have 3 channels (RGB), and the length of the shorter side is 1024. Approximately 18.7\% of all DHS clusters, spanning 48 countries, have a non-zero number of Mapillary images. Of these clusters with Mapillary images, the number of images ranges from 1 to a maximum of 300, with a mean of 76 and median of 94. The total number of Mapillary images included in \bench{} is approximately 1.7 million. \Cref{app:fig:mapillary} shows some example Mapillary images.

\paragraph{Comparison with Related Works}

\Cref{app:tab:dhs_compare} summarizes the related works for the DHS-based tasks in \bench.

As shown in \Cref{app:tab:compare_dhs_poverty}, the DHS-based datasets in \bench{} build on the previous works of \citealt{jean2016combining} and \citealt{yeh2020using}, which pioneered the application of computer vision on satellite imagery to estimate a cluster-level asset wealth index. Notably, for the task of predicting poverty over space, \bench{}'s dataset is nearly 5$\times$ larger than the dataset included in \cite{yeh2020using} (over 2$\times$ the number of countries, and 3$\times$ the temporal coverage). Our dataset also has advantages over other related works which often rely on proprietary imagery inputs \cite{jean2016combining,head2017can,gebru2017using}, are limited to a small number of countries \cite{babenko2017poverty,engstrom2017poverty,lee2021predicting,gebru2017using,watmough2019socioecologically}, or have coarser label resolution \cite{noor2008using}. Other researchers have explored using non-imagery inputs for poverty prediction, including Wikipedia text data \cite{sheehan2019predicting} and cell phone records \cite{blumenstock2015predicting}; while such multi-modal data are not currently in \bench, we are considering including them in future versions.

For the non-poverty tasks pertaining to health, education, and water/sanitation, there are extremely few ML-friendly datasets. \citealt{head2017can} comes closest to \bench{} in having predicted similar indicators (women BMI, women education, and clean water) derived from DHS survey data. Also, like us, their results suggest that satellite imagery may be less accurate at predicting these non-poverty labels in developing countries. However, because they used proprietary imagery inputs, their dataset is not accessible and cannot serve as a public benchmark. A large collaborative effort \cite{deshpande2020mapping} gathered survey and census data for creating clean water and sanitation labels in over 80 countries, but they did not provide satellite imagery inputs and only publicly released outputs of their geostatistical model, not the labels themselves. Again, \bench{} has significant advantages over other related works that use proprietary data \cite{head2017can,gebru2017using,maharana2018use}, are limited to a small number of countries \cite{gebru2017using,lee2021predicting}, or do not publicly release their labels \cite{deshpande2020mapping}.

\begingroup
\begin{table}
\centering
\caption{Comparison of related datasets using satellite images to predict DHS asset wealth index. *The clusters in \bench{} are a superset of the clusters included in \cite{yeh2020using} except for 2 clusters that had fewer than the minimum of 5 observations we required for inclusion in \bench.}
\label{app:tab:compare_dhs_poverty}
\renewcommand{\arraystretch}{1.2} %
\begin{tabular}{>{\raggedright\arraybackslash}p{0.16\textwidth} >{\raggedright\arraybackslash}p{0.23\textwidth} >{\raggedright\arraybackslash}p{0.24\textwidth} >{\raggedright\arraybackslash}p{0.25\textwidth}}
\toprule
    & \textbf{Jean et al. (2016)} \cite{jean2016combining}
    & \textbf{Yeh et al. (2020)} \cite{yeh2020using}
    & \textbf{\bench} \\
\midrule
Geographic range
    & 5 countries in Africa
    & 23 countries in Africa
    & 56 countries in 6 continents \\
Temporal range
    & 2010-2013
    & 2009-2016
    & 1996-2019 \\
Dataset size
    & 3,034 clusters
    & 19,669 clusters
    & 86,936 clusters* \\
Labels
    & asset wealth index with different asset variables in PCA for each country
    & asset wealth index with PCA pooled over 30 countries (a superset of the 23 countries with provided imagery)
    & asset wealth index with PCA pooled over all 56 countries \\
Daytime satellite imagery
    & $\sim$2.5m/px Google Static Maps daytime images, 3 bands, proprietary license
    & 30m/px resolution, 7 bands, Landsat 5/7/8 surface reflectance 3-year median composites (binned to either 2009-11, 2012-14, or 2015-17), some cloud masking
    & 30m/px resolution, 7 bands, Landsat 5/7/8 surface reflectance 3-year median composites (centered on survey year), improved cloud masking \\
Nightlights
    & $\sim$1km/px DMSP-OLS Nighttime Lights (uncalibrated), annual composite chosen to match survey year
    & (2009-2011) $\sim$1km/px DMSP-OLS Radiance-Calibrated Nighttime Lights, without inter-satellite calibration, 3-year composite; \newline (2012-2017) $\sim$500m/px VIIRS Stray Light Corrected Nighttime Day/Night Band, 3-year median composite of monthly images
    & (1996-2011) $\sim$1km/px DMSP-OLS Radiance-Calibrated Nighttime Lights, with inter-satellite calibration, annual composite chosen closest to survey year;\newline (2012-2019) $\sim$500m/px VIIRS Nighttime Day/Night Band (these are higher quality than the stray light corrected images), 3-year median composite of monthly images \\
\bottomrule
\end{tabular}
\end{table}
\endgroup

\begin{table}
\centering
\caption{Non-exhaustive comparison of related works and datasets for predicting DHS-based labels from satellite imagery, street-level imagery, or other non-survey inputs. ``None'' indicates that, to the best of our knowledge, we are not aware of existing works that predict the DHS label at scale. ``SB'' is short for \bench. (While \cite{deshpande2020mapping} uses survey data as inputs, they generate a prediction map including for locations where survey data were not available.)}
\label{app:tab:dhs_compare}
\begin{tabular}{>{\raggedright\arraybackslash}p{0.2\textwidth} >{\raggedright\arraybackslash}p{0.21\textwidth} >{\raggedright\arraybackslash}p{0.22\textwidth} >{\raggedright\arraybackslash}p{0.23\textwidth}}
\toprule
& \textbf{Satellite imagery} & \textbf{Street-level imagery} & \textbf{Other inputs} \\
\midrule
\textbf{poverty} \newline
SB includes 56 countries
    & \cite{jean2016combining} (5 countries) \newline
    \cite{yeh2020using} (23 countries) \newline
    \cite{noor2008using} (37 countries) \newline
    \cite{head2017can} (4 countries) \newline
    \cite{babenko2017poverty} (Mexico) \newline
    \cite{engstrom2017poverty} (Sri Lanka) \newline
    \cite{watmough2019socioecologically} (Kenya)
    & \cite{lee2021predicting} (2 countries) \newline
    \cite{gebru2017using} (USA)
    & \cite{sheehan2019predicting} (Wikipedia text, 31 countries) \newline
    \cite{blumenstock2015predicting} (phone records, Rwanda)
\\ \midrule
\textbf{women BMI} \newline
SB includes 53 countries
    & \cite{head2017can} (4 countries) \newline
    \cite{maharana2018use} (USA)
    & \cite{lee2021predicting} (India)
    & none
\\ \midrule
\textbf{child mortality} \newline
SB includes 56 countries
    & none
    & none
    & none
\\ \midrule
\textbf{women education} \newline
SB includes 56 countries
    & \cite{head2017can} (4 countries) \newline
    \cite{zhao2020framework} (9 countries)
    & \cite{gebru2017using} (USA)
    & none
\\ \midrule
\textbf{clean water} \newline
SB includes 49 countries
    & \cite{head2017can} (4 countries)
    & none
    & \cite{deshpande2020mapping} (survey data, 88 countries)
\\ \midrule
\textbf{sanitation} \newline
SB includes 49 countries
    & none
    & none
    & \cite{deshpande2020mapping} (survey data, 89 countries)
\\
\bottomrule
\end{tabular}
\end{table}

\paragraph{Dataset Impact}

Most low-income regions lack data on income and wealth at fine spatial scales. Even at coarse spatial scales, temporal resolution can still be bad; Figure 1 in \citealt{burke2021using} shows that, in some countries, as many as two decades can pass between successive nationally representative economic surveys. Inferring economic welfare from satellite or street-level imagery offers one solution to the lack of surveys.

Indeed, many governments turned to ML-based poverty mapping techniques during the COVID-19 pandemic to identify and prioritize vulnerable populations for targeted aid programs. For example, the government of Togo wanted to send aid to over 500,000 vulnerable people impacted by the pandemic. But like most low-income countries, Togo lacks accurate data on income and wealth at fine spatial scales. Working with a research group at UC Berkeley \cite{aiken2021machine,blumenstock2020machine}, the government was able to quickly deploy ML-based poverty mapping methods with satellite imagery inputs in order to identify who needs aid the most and then target cash payments to them. Likewise, the governments of Nigeria \cite{lowe2021national}, Mozambique, Liberia, and the Democratic Republic of the Congo \cite{gentilini2021cash} also used satellite imagery analysis for identifying and prioritizing neighborhoods with vulnerable individuals for their targeted social protection programs.

Finally, we highlight how ML-based poverty maps can feed into other policy evaluations. Researchers recently combined longitudinal ML-generated poverty maps of rural Uganda with data on expansion of the electric grid. By applying causal inference approaches, they were able to infer the impact of electrification on local livelihoods~\cite{ratledge2021using}. This work presents a scalable technique for measuring the effectiveness of large-scale infrastructure investments.

\subsection{Data for Predicting Change in Poverty Over Time}
\label{app:lsms}

\begin{figure}
\centering
    \begin{subfigure}[b]{0.4\textwidth}
    \centering
    \includegraphics[width=\textwidth]{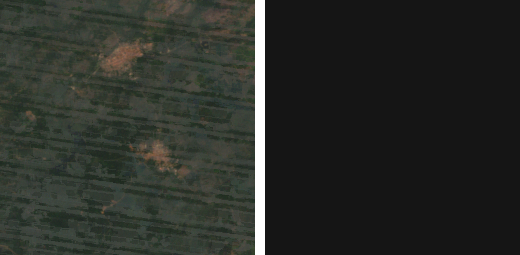}
    \caption{}
    \end{subfigure}
    \qquad
    \begin{subfigure}[b]{0.4\textwidth}
    \centering
    \includegraphics[width=\textwidth]{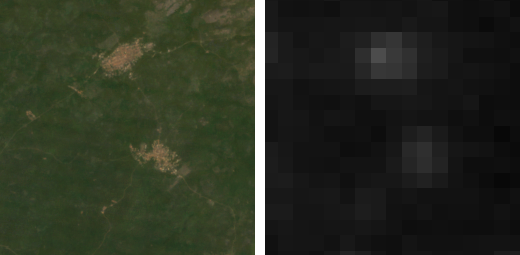}
    \caption{}
    \end{subfigure}
\caption{An example of a pair of satellite imagery inputs for predicting change in poverty over time for the Nigeria cluster located at (7.797380, 4.778803), in (a) 2010 and (b) 2015. Landsat RGB bands (left) and the DMSP/VIIRS nightlights band (right) are shown for each year.}
\label{app:fig:lsms_inputs}
\end{figure}

The task of predicting change in poverty over time uses labels calculated from household surveys conducted by the World Bank's Living Standards Measurement Study (LSMS) program. The LSMS surveys are similar to the DHS surveys described in the previous section. However, unlike DHS surveys, LSMS provides panel data---\emph{i.e.}, the same households are surveyed over time, facilitating comparison over time.

We start by compiling the same survey variables from the DHS asset index, except for refrigerator ownership because it is not included in the LSMS Uganda survey. (See the previous section for details on the survey variables included for the DHS asset index.) As with the DHS asset index, we convert ``floor type'', ``water source type'', and ``toilet type'' variables from descriptive categorical variables to a 1-5 ranked scale.

Based on the panel survey data, we calculate two PCA-based measures of change in asset wealth over time for each household: $\mathsf{diffOfIndex}$ and $\mathsf{indexOfDiff}$. For $\mathsf{diffOfIndex}$, we first assign each household-year an asset index computed as the first principal component of all the asset variables; this is the same approach used for the DHS asset index. Then, for each household, we calculate the difference in the asset index across years, which yields a ``change in asset index'' (hence the name $\mathsf{diffOfIndex}$). In contrast, $\mathsf{indexOfDiff}$ is created by first calculating the difference in asset variables in households across pairs of surveys for each country and then computing the first principal component of these differences; for each household, this yields a ``index of change in assets'' across years (hence the name $\mathsf{indexOfDiff}$). These measures are then averaged to the cluster-level to create cluster-level labels. We excluded a cluster if it contained fewer than 3 surveyed households.

As an example, consider an Ethiopian household $h$ that is surveyed in 2011 and 2015. This household would have 2 labels:
\begin{align*}
\mathsf{diffOfIndex}(h, 2011, 2015) &= \mathsf{assetIndex}(h, 2015) - \mathsf{assetIndex}(h, 2011) \\
\mathsf{indexOfDiff}(h, 2011, 2015) &= \mathsf{firstPrincipalComponent}(\mathsf{assets}(h, 2015) - \mathsf{assets}(h, 2011))
\end{align*}
If the set $\mathcal{C}$ of households represents a cluster in Ethiopia, then its cluster-level labels are
\begin{align*}
\mathsf{diffOfIndex}(\mathcal{C}, 2011, 2015) &= \frac{1}{|\mathcal{C}|} \sum_{h \in \mathcal{C}} \mathsf{diffOfIndex}(h, 2011, 2015) \\
\mathsf{indexOfDiff}(\mathcal{C}, 2011, 2015) &= \frac{1}{|\mathcal{C}|} \sum_{h \in \mathcal{C}} \mathsf{indexOfDiff}(h, 2011, 2015)
\end{align*}

The LSMS-based labels include data for 2,763 cluster-years (comprising 17,215 household-years) from 11 surveys for 5 African countries. \Cref{tab:lsms_surveys} gives the full list of LSMS surveys used,\footnote{LSMS survey data citations (all data was downloaded from \url{https://microdata.worldbank.org}):

Central Statistical Agency of Ethiopia. Ethiopia Rural Socioeconomic Survey (ERSS) 2011-2012. Public Use Dataset. Ref: \verb|ETH_2011_ERSS_v02_M|. Downloaded on August 25, 2021.

Central Statistical Agency of Ethiopia. Ethiopia Socioeconomic Survey, Wave 3 (ESS3) 2015-2016. Public Use Dataset. Ref: \verb|ETH_2015_ESS_v02_M|. Downloaded on August 26, 2021.

National Statistical Office, Government of Malawi. Integrated Household Panel Survey (IHPS) 2010-2013-2016. Public Use Dataset. Ref: \verb|MWI_2010-2016_IHPS_v03_M|. Downloaded on September 3, 2021.

National Bureau of Statistics, Federal Republic of Nigeria. Nigeria General Household Survey (GHS), Panel 2010, Wave 1. Ref: \verb|NGA_2010_GHSP-W1_v03_M|. Dataset downloaded on September 4, 2021.

National Bureau of Statistics, Federal Republic of Nigeria. Nigeria General Household Survey (GHS), Panel 2015-2016, Wave 3. Ref: \verb|NGA_2015_GHSP-W3_v02_M|. Dataset downloaded on September 4, 2021.

Tanzania National Bureau of Statistics (NBS). Tanzania National Panel Survey 2008-2009 (Round 1). Ref: \verb|TZA_2008_NPS-R1_v03_M|. Dataset downloaded on September 4, 2021.

Tanzania National Bureau of Statistics (NBS). Tanzania National Panel Survey Report (NPS) - Wave 2, 2010-2011. Dar es Salaam, Tanzania: NBS. Ref: \verb|TZA_2010_NPS-R2_v03_M|. Dataset downloaded on September 5, 2021.

Tanzania National Bureau of Statistics (NBS). Tanzania National Panel Survey Report (NPS) - Wave 3, 2012-2013. Dar es Salaam, Tanzania: NBS. Ref: \verb|TZA_2012_NPS-R3_v01_M|. Dataset downloaded on September 4, 2021.

Uganda Bureau of Statistics. Uganda National Panel Survey (UNPS), 2005-2009. Public Use Dataset. Ref: \verb|UGA_2005-2009_UNPS_v01_M|. Downloaded on August 25, 2021.

Uganda Bureau of Statistics. Uganda National Panel Survey (UNPS), 2013-2014. Public Use Dataset. Ref: \verb|UGA_2013_UNPS_v01_M|. Downloaded on August 25, 2021.} and \Cref{tab:lsms_counts} gives the number of clusters and households included for each country. See \Cref{app:fig:lsms_inputs} for an example of the satellite imagery inputs.

The labels and inputs provided in \bench{} for this task are similar (but not identical) to the labels and inputs used in \cite{yeh2020using}. While the underlying LSMS survey data used are the same, there are 3 key differences.
\begin{enumerate}
    \item In \bench, for each country, we only used data from households that are present in all surveys of that country. In Uganda, for example, we only keep households that were surveyed repeatedly in all of the 2005, 2009, and 2013 surveys. This is different from \cite{yeh2020using} which included any household that was present in two survey years---\emph{e.g.}, a household in Uganda 2005 and Uganda 2009, but not Uganda 2013.

    \item The recoding of the floor, water, and toilet quality variables was made more consistent across countries and now closely matches the ranking introduced in \cite{deshpande2020mapping}.

    \item As in the case of the DHS-based datasets, the satellite imagery inputs have been improved. See \Cref{app:tab:compare_dhs_poverty} for details.
\end{enumerate}

\begin{table}
\centering
\caption{LSMS surveys}
\label{tab:lsms_surveys}
\begin{adjustbox}{max width=0.99\linewidth}
\begin{tabular}{ll>{\scriptsize\ttfamily}l}
    \toprule
    Country and Year & Survey Title & \footnotesize{\textrm{Survey ID}} \\
    \midrule
    Ethiopia 2011       & Rural Socioeconomic Survey 2011-2012              & ETH\_2011\_ERSS\_v02\_M \\
    Ethiopia 2015       & Socioeconomic Survey 2015-2016, Wave 3            & ETH\_2015\_ESS\_v03\_M \\
    Malawi 2010 \& 2016 & Integrated Household Panel Survey 2010-2013-2016  & MWI\_2010-2016\_IHPS\_v03\_M \\
    Nigeria 2010        & General Household Survey, Panel 2010-2011, Wave 1 & NGA\_2010\_GHSP-W1\_V03\_M \\
    Nigeria 2015        & General Household Survey, Panel 2015-2016, Wave 3 & NGA\_2015\_GHSP-W3\_v02\_M \\
    Tanzania 2008       & National Panel Survey 2008-2009, Wave 1           & TZA\_2008\_NPS-R1\_v03\_M \\
    Tanzania 2012       & National Panel Survey 2012-2013, Wave 3           & TZA\_2012\_NPS-R3\_v01\_M \\
    Uganda 2005 \& 2009 & National Panel Survey 2005-2009                   & UGA\_2005-2009\_UNPS\_v01\_M \\
    Uganda 2013         & National Panel Survey 2013-2014                   & UGA\_2013\_UNPS\_v01\_M \\
    \bottomrule
\end{tabular}
\end{adjustbox}
\end{table}

\begin{table}
\centering
\caption{Number of clusters and households included from each country for the ``predicting change in poverty over time'' task, based on LSMS survey data.}
\label{tab:lsms_counts}
\begin{tabular}{l c c}
\toprule
Country  & \# clusters & \# households \\
\midrule
Ethiopia &         235 &          1128 \\
Malawi   &         101 &          1085 \\
Nigeria  &         462 &          3093 \\
Tanzania &         300 &          1431 \\
Uganda   &         189 &          1247 \\
\midrule
Total    &        1287 &          7984 \\
\bottomrule
\end{tabular}
\end{table}

\paragraph{Comparison with Related Works}
To the best of our knowledge, the LSMS-based poverty change over time dataset in \bench{} and its predecessor in \cite{yeh2020using} are the only datasets specifically designed as an index of asset wealth change. For related works on mapping poverty, see the ``Comparison with Related Works'' for DHS-based tasks in \Cref{app:dhs}.

\subsection{Cropland Mapping with Landsat}
\label{app:sec:crop_mapping}

\begin{figure}
    \centering
    \includegraphics[width=0.99\textwidth]{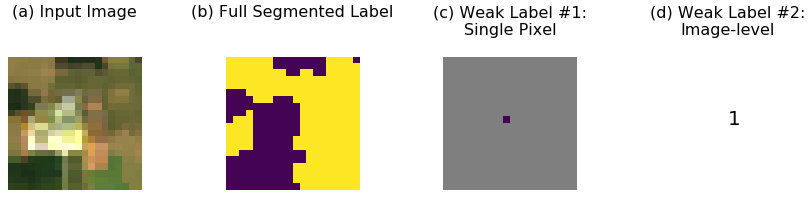}
    \caption{An example from the cropland mapping dataset \cite{wang2020weakly}, showing (a) an example Landsat image, (b) its corresponding fully segmented label, (c) single pixel weak label, and (d) image-level weak label.}
    \label{app:fig:cropland_mapping}
\end{figure}

We release a dataset for performing weakly supervised classification of cropland in the United States using the data from \citealt{wang2020weakly}, which has not been released previously.
While densely segmented labels are time-consuming and infeasible to generate for a region as large as Sub-Saharan Africa, pixel-level and image-level labels are often already available and much easier to create. \Cref{app:fig:cropland_mapping} shows an example from the dataset.

The study area spans from 37$^{\circ}$N to 41$^{\circ}$30'N and from 94$^{\circ}$W to 86$^{\circ}$W, and covers an area of over 450,000km$^{2}$ in the United States Midwest. We chose this region because the US Department of Agriculture (USDA) maintains high-quality pixel-level land cover labels across the US \cite{cdl}, allowing us to evaluate the performance of algorithms. Land cover-wise, the study region is 44\% cropland and 56\% non-crop (mostly temperate forest).

The Landsat Program is a series of Earth-observing satellites jointly managed by the USGS and NASA. Landsat 8 provides moderate-resolution (30m) satellite imagery in seven surface reflectance bands (ultra blue, blue, green, red, near infrared, shortwave infrared 1, shortwave infrared 2) designed to serve a wide range of scientific applications. Images are collected on a 16-day cycle.

We computed a single composite by taking the median value at each pixel and band from January 1, 2017 to December 31, 2017. We used the quality assessment band delivered with the Landsat 8 images to mask out clouds and shadows prior to computing the median composite. The resulting seven-band image spans 4.5 degrees latitude and 8.0 degrees longitude and contains just over 500 million pixels. The composite was then divided into 200,000 tiles of $50\times 50$ pixels each. This full dataset was not released previously with \citealt{wang2020weakly}.

The ground truth labels from the Cropland Data Layer \cite{cdl} are at the same spatial resolution as Landsat, so that for every Landsat pixel there is a corresponding $\{\text{cropland}, \text{not cropland}\}$ label. For each image, we generate two types of weak labels: (1) single pixel and (2) image-level, both with the goal of generating dense semantic segmentation predictions. The image-level label is $\in \{\ge 50\% \text{ cropland}, < 50\% \text{ cropland}\}$.

\paragraph{Comparison with Related Works}

Cropland has already been mapped globally \cite{buchhorn2020copernicus, friedl2002global} or for the continent of Africa \cite{xiong2017automated} in multiple state-of-the-art land cover maps. However, existing land cover maps are known to have low accuracy throughout the Global South \cite{kerner2020rapid}. One reason behind this low accuracy is that existing maps have been created with SVM or tree-based algorithms that take into account a single pixel at a time \cite{buchhorn2020copernicus, friedl2002global, xiong2017automated}. \citealt{kerner2020rapid} showed that a multi-headed LSTM (still trained on single pixels) outperformed SVM and random forest classifiers on cropland prediction in Togo. Using a larger spatial context, \emph{e.g.}, in a CNN, could lead to further accuracy gains. However, ground label scarcity remains a bottleneck for applying deep learning models to map cropland. \citealt{wang2020weakly} showed that weak labels in the form of single pixel or image-level classes can still supervise a U-Net to segment cropland at accuracies better than SVM or random forest classifiers. We release this dataset, which is the first dataset for weakly supervised cropland mapping, as a benchmark for algorithm development. The dataset is in the U.S. Midwest because cropland labels there are of high accuracy; methods developed on this dataset could be paired with newly generated weak labels in low-income regions to generate novel, high-accuracy cropland maps (see below for an example application).

\paragraph{Dataset Impact}

High accuracy cropland mapping in the Global South can have significant impacts on the planning of government programs and downstream tasks like crop type mapping and yield prediction. For instance, during the COVID-19 pandemic, the government of Togo announced a program to boost national food production by distributing aid to farmers. However, the government lacked high-resolution spatial information about the distribution of farms across Togo, which was crucial for designing this program. Existing global land cover maps, despite including a cropland class, were low in accuracy across Togo. The government collaborated with researchers at the University of Maryland to solve this problem, and in \citealt{kerner2020rapid} the authors created a high-resolution map of cropland in Togo for 2019 in under 10 days. The authors pointed out that this case study demonstrates ``a successful transition of machine learning research to operational rapid response for a real humanitarian crisis'' \cite{kerner2020rapid}.

\subsection{Crop Type Mapping with Planet and Sentinel Imagery}
\label{app:sec:croptype1}

\begin{figure}
    \centering
    \includegraphics[width=0.3\textwidth]{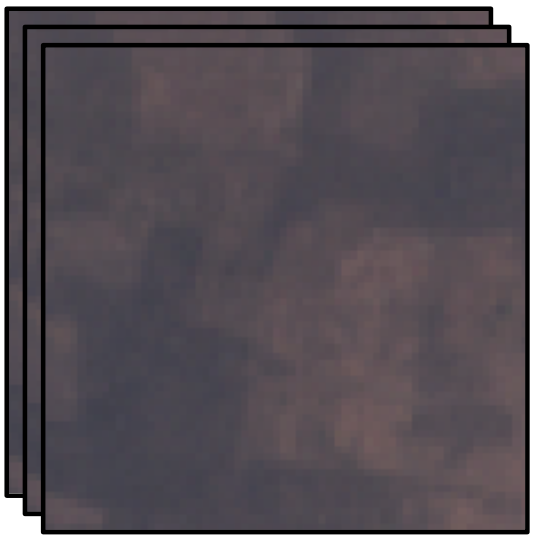}
    \qquad
    \includegraphics[width=0.3\textwidth]{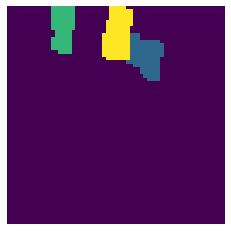}
    \caption{An example from the crop type mapping dataset \cite{rustowicz2019semantic}. The left image represents a satellite image timeseries (figure displays PlanetScope imagery) and the right image represents a segmentation map.}
    \label{app:fig:croptype_mapping}
\end{figure}
As introduced in \cite{rustowicz2019semantic}, these datasets contain satellite imagery from Ghana and South Sudan. Sentinel 1 (10m resolution), Sentinel 2 (10m resolution), and Planet's PlanetScope (3m resolution) time series imagery are used as inputs for this task. As described in \cite{rustowicz2019semantic}, Planet imagery is incorporated to help mitigate issues from high cloud cover and small field sizes. We include three S1 bands (VV, VH, VH/VV), ten S2 bands (blue, green, red, near infrared, four red edge bands, two short wave infrared bands), and all four PlanetScope bands (blue, green, red, near infrared). We also construct normalized difference vegetation index (NDVI) and green chlorophyll vegetation index (GCVI) bands for PlanetScope and S2 imagery.

Ground truth labels consist of a 64x64 pixel segmentation map, with each pixel containing a crop label. Ghana locations are labeled for Maize, Groundnut, Rice, and Soya Bean, while South Sudan locations are labeled for Sorghum, Maize, Rice, and Groundnut.

\begin{figure}
    \centering
    \includegraphics[width=0.3\textwidth]{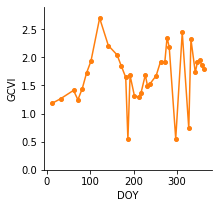}
    \qquad
    \includegraphics[width=0.3\textwidth]{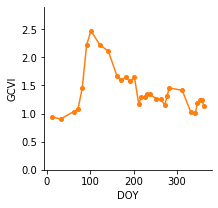}
    \caption{Example time series of the GCVI band computed from Sentinel-2 satellite bands \cite{kluger2021two}, after clouds were masked out. Both examples happen to be of the crop type ``Cassava''.}
    \label{app:fig:croptype_mapping_kenya}
\end{figure}

\paragraph{Comparison with Related Works}

\begin{table}[tbp]
\centering
\caption{\added{A comparison of \bench{}'s crop type datasets with existing datasets. A dataset is only included if it is designed for crop type mapping, is publicly available, and provides both inputs and outputs in ML-friendly formats. Compared to \Cref{tab:compare_benchmarks}, we include datasets that lack train/test splits and standardized benchmarks, though we make a note of their existence in the columns.}}
\label{app:tab:croptype_data}
\begin{adjustbox}{max width=0.99\linewidth}
\begin{tabular}{>{\raggedright\arraybackslash}p{3cm} 
>{\raggedright\arraybackslash}p{1.7cm}
>{\raggedright\arraybackslash}p{2cm} >{\raggedright\arraybackslash}p{2cm} >{\raggedright\arraybackslash}p{2cm} >{\raggedright\arraybackslash}p{2cm}
>{\centering\arraybackslash}p{1cm}
>{\centering\arraybackslash}p{1cm}
>{\centering\arraybackslash}p{1.2cm}}
\toprule
\textbf{Dataset Collection}
    & \textbf{Dataset \newline \#}
    & \textbf{Geography} 
    & \textbf{Time} 
    & \textbf{Inputs} 
    & \textbf{Size}
    & \textbf{Small-\newline holder?}
    & \textbf{Data \newline splits?}
    & \textbf{Base- \newline line?}
    \\
\midrule
\multirow{4}{*}{\bench{}}
    & 1 \cite{rustowicz2019semantic}
    & Ghana and South Sudan
    & 2016-17
    & Sat. image time series
    & 4,439 and 837 fields
    & \checkmark
    & \checkmark
    & \checkmark
    \\
\cmidrule{2-9}
    & 2 \cite{jin2019smallholder, kluger2021two}
    & Kenya
    & 2017
    & Sat. time series
    & 5,746 fields
    & \checkmark
    & \checkmark
    & \checkmark
    \\
    \midrule
\multirow{16}{*}{Radiant MLHub \cite{radiant}}
    & 1 \cite{rustowicz2019semantic}
    & Ghana and South Sudan
    & 2016-17
    & Sat. image time series
    & 4,439 and 837 fields
    & \checkmark
    & \checkmark
    & \checkmark
    \\
\cmidrule{2-9}
    & 2 \cite{kerner2020fieldlevel}
    & Kenya
    & 2019
    & Sat. image time series
    & 4,668 fields
    & \checkmark
    & \checkmark
    & \checkmark
    \\
\cmidrule{2-9}
    & 3
    & Kenya
    & 2019
    & Sat. image time series
    & 319 fields
    & \checkmark
    \\
\cmidrule{2-9}
    & 4
    & Tanzania
    & 2019
    & Sat. image time series
    & 392 fields
    & \checkmark
    \\
\cmidrule{2-9}
    & 5
    & Uganda
    & 2017
    & Sat. image time series
    & 232 fields
    & \checkmark
    \\
\cmidrule{2-9}
    & 6 \cite{chew2020deep}
    & Rwanda
    & 2018-19
    & Drone imagery
    & 2,611 points
    & \checkmark
    & \checkmark
    & \checkmark
    \\
\cmidrule{2-9}
    & 7 \cite{remelgado2020a}
    & Uzbekistan and Tajikistan
    & 2015-18
    & Sat. imagery time series
    & 8,435 fields
    &
    \\
\cmidrule{2-9}
    & 8
    & South Africa
    & 2017-18
    & Sat. imagery time series
    & Unknown
    &
    & \checkmark
    & \checkmark
    \\
\bottomrule
\end{tabular}
\end{adjustbox}
\end{table}

\bench{}'s crop type datasets and existing crop type datasets are summarized in \Cref{app:tab:croptype_data}. A version of \bench{}'s Ghana/South Sudan dataset was released previously and is currently housed on Radiant MLHub. We highlight key differences between \bench{}'s dataset and the one used in \citealt{rustowicz2019semantic}. We use the same train, validation, and test splits used in \cite{rustowicz2019semantic}, though we use the full 64x64 imagery provided, while \cite{rustowicz2019semantic} further subdivided imagery into 32x32 pixel grids due to memory constraints. We also include variable length time series with zero padding and masking, while \cite{rustowicz2019semantic} trimmed the respective time series down to the same length. We include variable length time series with the reasoning that future research should be extendable to variable length time-series imagery. The metrics cited in \Cref{tab:benchmark} are on the original \citealt{rustowicz2019semantic} dataset.

\subsection{Crop Type Mapping with Sentinel-2 Time Series}
\label{app:sec:croptype2}

The data from \citealt{jin2019smallholder} and \citealt{kluger2021two} comes from three regions in Kenya: Bungoma, Busia, and Siaya. They use time series from the multi-spectral Sentinel-2 (10m resolution) to differentiate crop types at individual pixels in the fields (\Cref{app:fig:croptype_mapping_kenya}). Time series span from January 1, 2017 to December 31, 2017. All 13 Sentinel-2 bands were used as features, along with GCVI (green chlorophyll vegetation index) as a fourteenth band. Cloudy observations were removed using the QA60 band delivered with Sentinel-2 and the Hollstein Quality Assessment measure.

Ground truth labels are from a survey conducted on crop types during the long rains season in Kenya in 2017. The labels span 9 crop types: Sweet Potatoes, Cassava, Maize, Banana, Beans, Groundnut, Sugar Cane, Other, and Non-crop.

The train, validation, and test sets are split by region to encourage discovery of features and development of methods that generalize across regions. One region is the training and validation region, while the other two regions are test regions.

\paragraph{Comparison with Related Works}

\bench{}'s crop type datasets and existing crop type datasets are compared in \Cref{app:tab:croptype_data}. A dataset was only included in the table if it is publicly available and provides inputs and outputs in ML-friendly formats. There is considerable work underway in the remote sensing community, led by the Radiant Earth Foundation, to collect and disperse crop type data to improve the state-of-the-art classification. \bench{}'s crop type dataset in Kenya complements existing datasets. It is one of the largest available crop type datasets in a smallholder system. It also has defined train/val/test splits and baselines, which not all public crop type datasets do. One of the train/val/test split options is also designed to test model generalizability across geography by splitting along geographic clusters, which no other datasets do. We recommend that ML researchers test their methods on as many available datasets as possible to ensure model generalizability.

\paragraph{Dataset Impact}
The crop type labels that we released in Kenya were the same labels used to create the first-ever maize classification and yield map across that entire country \cite{jin2019smallholder}. Kenya is one of the largest maize producers in sub-Saharan Africa, and studying maize production there could improve food security in the region. \citealt{jin2019smallholder} used a random forest trained on seasonal median composites of satellite imagery to predict maize with an accuracy of only 63\%. It is worth investigating how other machine learning models using a year’s full time series could improve on this. As an example of novel insights resulting from one of our datasets: analysis of the maize yield map in \citealt{jin2019smallholder} revealed that 72\% of variation in predicted maize yields could be explained by soil factors, suggesting that increasing nitrogen fertilizer application should be a priority for increasing smallholder yields in Kenya.

\subsection{Crop Yields and MODIS} \label{app:sec:crop}

\begin{figure}
    \centering
    \includegraphics[width=0.3\textwidth]{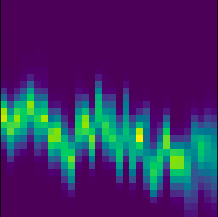}
    \qquad
    \includegraphics[width=0.3\textwidth]{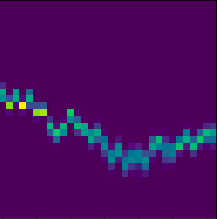}
    \caption{An example from the MODIS crop yield dataset. The spectral histograms are over the 2015 harvest season in the La Capital department, Santa Fe province, Argentina, with a soybean yield of 2.947 metric tonnes per hectare. The left image shows surface reflectance band 5 out of 7, covering wavelengths from 1230-1250nm. The right image shows surface temperature band 1, covering daytime land surface temperatures.}
    \label{app:fig:modis}
\end{figure}

These datasets are constructed as an expansion of the dataset used in \cite{wang2018transfer}. They are created using Moderate Resolution Imaging Spectroradiometer (MODIS) satellite imagery, which is freely accessible via Google Earth Engine and provides coverage of the entire globe. Specifically, we use 8-day composites of MODIS images to get 7 bands of surface reflectance at different wavelengths (3 visible and 4 infrared bands) from the MOD09A1 \cite{MODIS-MOD09A1} collection, 2 bands of day and night surface temperatures from MYD11A2 \cite{MODIS-MYD11A2}, and a land cover mask from MCD12Q1 \cite{MODIS-MCD12Q1} to distinguish cropland from other land. For each of the 9 bands of reflectance and temperature imagery and each of the 32 timesteps within a year's harvest season, we bin pixel values into 32 ranges, giving a $32 \times 32 \times 9$ final histogram. We create one such dataset for each of Argentina, Brazil, and the United States, with 9049 datapoints for the United States, 1615 for Argentina, and 384 for Brazil.
\\ \\
The ground truth labels are the regional crop yield per harvest, in metric tonnes per cultivated hectare, as collected from Argentine Undersecretary of Agriculture \cite{cropyield-argentina}, the Brazilian Institute of Geography and Statistics \cite{cropyield-brazil}, and the United States Department of Agriculture \cite{cropyield-usa}.

\paragraph{Comparison with Related Works}

\bench{} releases the crop yield datasets from two previous works \cite{you2017deep,wang2018transfer} for the first time. To date, very few crop yield datasets exist, because yields require expensive farm survey techniques (\emph{e.g.}, crop cuts) to measure. The datasets that do contain field-level yields are privately held by researchers, government agencies, or NGOs. \bench{}'s datasets therefore provide yields at the county level. Furthermore, crop yield prediction is challenging as it requires processing a temporal sequence of satellite images. We provide ML-friendly inputs in the form of histograms of weather and satellite features over each county.

\paragraph{Dataset Impact}

Tracking crop yields is crucial to measuring agricultural development and deciding resource allocation, with downstream applications to food security, price stability, and agricultural worker income. Notably, most developed countries invest in forecasting and tracking crop yield. For example, the European Commission JRC's crop yield forecasts and crop production estimates inform the EU's Common Agricultural Policy and other agricultural programs \cite{eu-cropforecasting}. By involving satellite images in the crop yield prediction process, we aim to make timely predictions available in developing countries where ground surveys are costly and infrequent. Furthermore, we provide satellite histograms rather than human-engineered indices like NDVI, which are more human-friendly for visualization but discard a significant amount of potentially-relevant information. In doing so, we hope to encourage the development of ML techniques that make use of more complete and useful features to generate better predictions.

\subsection{Field Delineation with Sentinel-2}

\begin{figure}
     \centering
     \begin{subfigure}[b]{0.3\textwidth}
         \centering
         \includegraphics[width=\textwidth]{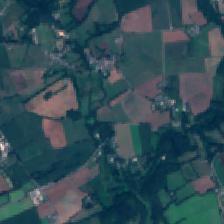}
         \caption{Sentinel-2 (Input)}
     \end{subfigure}
     \hfill
     \begin{subfigure}[b]{0.3\textwidth}
         \centering
         \includegraphics[width=\textwidth]{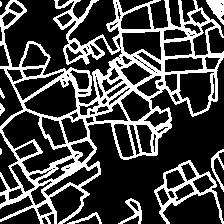}
         \caption{Delineated boundaries}
     \end{subfigure}
     \hfill
     \begin{subfigure}[b]{0.3\textwidth}
         \centering
         \includegraphics[width=\textwidth]{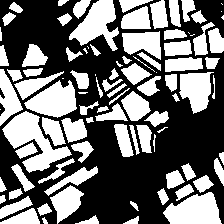}
         \caption{Segmentation masks}
     \end{subfigure}
        \caption{An example from the field delineation dataset~\cite{aung2020farm}. From left to right, an input Sentinel-2 image, its corresponding delineated boundaries, and its corresponding segmentation masks.}
        \label{app:fig:farmland}
\end{figure}

As introduced in \cite{aung2020farm}, the dataset consists of Sentinel-2 satellite imagery in France\footnote{\url{https://www.data.gouv.fr/en/datasets/registre-parcellaire-graphique-rpg-contours-des-parcelles-et-ilots-culturaux-et-leur-groupe-de-cultures-majoritaire/}} over the 3 time ranges January-March, April-June, and July-September in 2017. The image has resolution $224\times 224$ corresponding to a 2.24km$\times$ 2.24km area on the ground. Each satellite image comes along with the corresponding binary masks of boundaries and areas of farm parcels.
The dataset consists of 1572 training samples, 198 validation samples, and 196 test samples. We use a different data split from \citealt{aung2020farm} to remove overlapping between the train, validation and test split.
An example of the dataset is shown in \Cref{app:fig:farmland}.

\paragraph{Comparison with Related Works}

To our knowledge, SustainBench has released the first public field boundary dataset with satellite image inputs and ML-friendly outputs. That is, some countries in Europe (\emph{e.g.}, France) have made vector files of field boundaries public on their government websites, but without corresponding satellite imagery inputs or raster field boundary outputs. We provide these inputs and outputs. While field segmentation datasets from the U.S., South Africa, and Australia were used in prior field delineation research \cite{yan2016conterminous, waldner2020deep, waldner2021detect}, none of those datasets are publicly available. We are also currently working on collecting field boundaries in low-income countries, but this data will be added to SustainBench at a later date, not in time for this submission.

\paragraph{Dataset Impact}

Automated field delineation makes it easier for farmers to access field-level analytics; previously, manual boundary input was a major deterrent from adopting digital agriculture \cite{waldner2020deep}. Digital agriculture can improve yields while minimizing the use of inputs like fertilizer that cause environmental pollution -- with the net effect of increasing farmer profit. The development of a new attention-based neural network architecture (called FracTAL ResUNet) enabled the delineation of 1.7 million fields in Australia from satellite imagery \cite{waldner2021detect}. These field boundaries have since been productized  by CSIRO, the Australian government agency for scientific research.  This is an example where a novel deep learning architecture enabled the creation of operational products in agriculture. However, the Australia dataset is not publicly available. Our goal is for the release of \bench{}'s field boundary dataset in France to enable further architecture development and identify which model works best for field delineation.

\subsection{Brick Kiln Detection with Sentinel-2}
\label{app:remote_sensing}

Brick manufacturing is a major source of pollution in South Asia, but the industry is largely comprised of small-scale, informal producers, making it difficult to monitor and regulate. Identifying brick kilns automatically from satellite imagery can help improve compliance with environmental regulations and measure their impact on the health of nearby populations. We provide Sentinel-2 satellite imagery at 10m/pixel resolution available through Google Earth Engine~\cite{gorelick2017google}. The images have size 64$\times$64$\times $13px, where the order of the bands correspond to the bands B1 through B12 on the Earth Engine Data Catalog, where B2 is Blue, B3 is Green, and B4 is Red. The other bands include aerosols, color infrared, short-wave infrared, and water vapor data.

\paragraph{Comparison with Related Works}
A recent study detected brick kilns from high-resolution (1m/pixel) satellite imagery and hand-validated the predictions, providing ground truth locations of brick kilns in Bangladesh for the time period of October 2018 to May 2019 \cite{lee2021scalable}. The imagery could not be shared publicly because they were proprietary. Hence, we provide Sentinel-2 satellite imagery instead. With help from domain experts, we verified the labels of each image as not containing a brick kiln (class 0) or containing a brick kiln (class 1) based on the ground truth locations provided by \cite{lee2021scalable}. There were roughly 374,000 examples total, with 6329 positives. We sampled 25\% of the remaining negatives, removed any null values, and included the remaining 67,284 negative examples in our dataset.

\paragraph{Dataset Impact}
\bench{} introduces the first publicly released dataset of this size and quality on detecting brick kilns across Bangladesh from satellite imagery. This dataset was manually labeled and verified in-house by domain experts. Brick kiln detection is a challenging task because of the sparsity of kilns and lack of similar training data, but with recent developments in satellite monitoring \cite{lee2021scalable}, it plays a key role in affecting policy developed by public health experts, industry stakeholders (\emph{e.g.}, kiln owners), and government agencies \cite{stanfordwoodsbrick}. \bench{} is the first to contribute a large dataset for this task, and the results of models will be utilized by policymakers.

\begin{figure}
    \centering
    \includegraphics[width=0.3\textwidth]{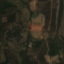}
    \qquad
    \includegraphics[width=0.3\textwidth]{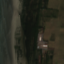}
    \caption{An example of Sentinel-2 satellite imagery for brick kiln classification. On the left is a positive example of an image showing a brick kiln, while the right image is a negative example (\emph{i.e.}, no brick kiln).}
    \label{app:fig:brick_kiln}
\end{figure}

\subsection{Representation Learning for Land Cover Classification}
\label{app:sec:tile2vec}

The dataset from \citealt{jean2019tile2vec} uses imagery from the USDA's National Agriculture Imagery Program (NAIP), which provides aerial imagery for public use that has four spectral bands (red (R), green
(G), blue (B), and infrared (N)) at 0.6 m ground resolution. They obtained an image of Central Valley, California near the city of Fresno for the year 2016, spanning latitudes [36.45, 37.05] and longitudes [-120.25, -119.65]. There are over 12 billion pixels in the dataset.

The Cropland Data Layer (CDL) is a raster georeferenced land cover map collected by the USDA for the continental United States \cite{cdl} and serves as ground truth labels of land cover. Offered at 30 m resolution, CDL includes 132 class labels spanning crops, developed areas, forest, water, and more. In the NAIP dataset over Central Valley, CA, 66 CDL classes are observed.  CDL is used as ground truth for evaluation by upsampling it to NAIP resolution and taking the mode over each NAIP image.

\begin{figure}
    \centering
    \includegraphics[width=0.50\textwidth]{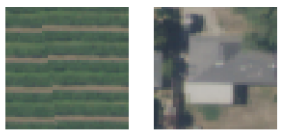}
    \caption{Example images from the NAIP dataset collected by \citealt{jean2019tile2vec}. The left image is an example of the ``Grapes'' class and the right image is an example of the ``Urban'' class.}
    \label{app:fig:tile2vec}
\end{figure}

\paragraph{Comparison with Related Works}

Representation learning on natural images often uses canonical computer vision datasets like ImageNet and Pascal VOC to evaluate new methods. Satellite imagery lacks an analogous dataset. The high-resolution aerial imagery dataset released in \bench{} aims to fill this void for land cover mapping with high-resolution inputs in particular. We note that, for object detection or lower resolution inputs, repurposing a dataset like fMoW \cite{christie2018functional}, SpaceNet \cite{vanetten2019spacenet}, Sen12MS \cite{schmitt2019sen12ms}, or BigEarthNet \cite{sumbul2019bigearthnet} would also be appropriate. To our knowledge, such repurposing has not yet been done.

\paragraph{Dataset Impact}

Many tasks in sustainability monitoring have abundant unlabeled imagery but scarce labels. Land cover mapping in low-income regions is one example; crop type mapping in smallholder systems is another. By learning representations of satellite images in an unsupervised or self-supervised way, we may be able to improve performance on SDG-related tasks for the same number of training labels.

\subsection{Out-of-Domain Land Cover Classification}
\label{app:sec:meta}

\citealt{wang2020meta} sampled one thousand $10$km $\times 10$km regions uniformly at random from the Earth's land surface, and removed regions that have fewer than 2 unique land cover classes and regions where one land cover type comprises more than 80\% of the region's area. This resulted in 692 regions across 105 countries. The authors placed the 103 regions from Sub-Saharan Africa into the meta-test set and split the remainder into 485 meta-train and 104 meta-val regions at random. We provide the user with the option of placing any continent into the meta-test set and splitting the other continents' regions at random between the meta-train and meta-val sets.

\begin{figure}
    \centering
    \includegraphics[width=0.3\textwidth]{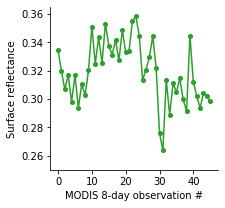}
    \qquad
    \includegraphics[width=0.3\textwidth]{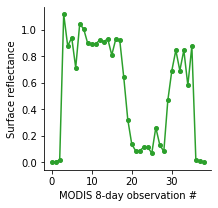}
    \caption{Example time series from the 8-day MODIS satellite product collected by \citealt{wang2020meta}. The left time series is an example from Mauritania and the right time series is an example from Canada.}
    \label{app:fig:metalearning}
\end{figure}

In each region, 500 points were sampled uniformly at random. At each point, the MODIS Terra Surface Reflectance 8-Day time series was exported for January 1, 2018 to December 31, 2018 (\Cref{app:fig:metalearning}). MODIS collects 7 bands and NDVI was computed as an eighth feature, resulting in a time series of dimension $8 \times 46$. Global land cover labels came from the MODIS Terra+Aqua Combined Land Cover Product, which classifies every 500m-by-500m pixel into one of 17 land cover classes (\emph{e.g.}, grassland, cropland, desert).

\paragraph{Comparison with Related Works}

This \bench{} dataset from \cite{wang2020meta} is the first time that any few-shot learning dataset has been released for satellite data. Because land cover products are available globally (albeit with varying accuracy), \citealt{wang2020meta} created a few-shot dataset for land cover classification.

\paragraph{Dataset Impact}

Our hope is that this dataset can be included in evaluations of few-shot learning algorithms to see how they do on real-world time series, and that new algorithms will improve knowledge sharing from high-income regions to low-income ones. That way, performance on remote sensing tasks can be increased in low-income regions for tasks with few labels.

\section{Benchmark Details}
\label{app:baseline_models}

Code to reproduce baseline models new to SustainBench can be found in our GitHub repo.

\subsection{DHS survey-based regression tasks (SDGs 1, 3, 4, 6)}

The DHS survey-based regression tasks include predicting an asset wealth index (SDG 1), women's BMI and child mortality rates (SDG 3), women's educational attainment (SDG 4), and water and sanitation indices (SDG 6). We adapt the \texttt{KNN scalar NL} model from \cite{yeh2020using} as the \bench{} baseline model for these tasks. We chose this model for its simplicity and its high performance on predicting asset wealth as noted in \cite{yeh2020using}. For each label, we fitted a $k$-nearest neighbor ($k$-NN) regressor implemented using scikit-learn, and the $k$ hyperparameter was tuned on the validation split, taking on integer values between 1 and 20, inclusive. The input to the $k$-NN model is the mean nightlights value from the nightlights band in the satellite input image, with separate models trained for the DMSP (survey year $\leq 2011$) vs. VIIRS (survey year $\geq$ 2012) bands.

\paragraph{Comparison with Related Works}

We observe that our KNN nightlights baseline model roughly matches the performance described in \cite{yeh2020using} on the poverty prediction over space task ($r^2$ = 0.63). However, its $r^2$ values for predicting the other non-poverty labels is much lower: child mortality rate ($r^2$ = 0.01), women BMI (0.42), women education (0.26), water index (0.40), sanitation index (0.36). Our result is in line with a similar observation made by \cite{head2017can}, which also found that models trained on satellite images were better at predicting the asset wealth index than other non-poverty labels in 4 African countries. This strongly suggests that predicting these other labels almost certainly requires different models and/or inputs. Indeed, this is why \bench{} provides street-level imagery in addition to satellite imagery.

While \bench{} also provides street-level images for many DHS clusters, we do not have any baseline models yet that take advantage of the street-level imagery. Some preliminary results using street-level imagery to predict asset wealth and women's BMI are shown in \cite{lee2021predicting}, although they only tested their models on India and Kenya (compared to the $\sim$50 countries included for DHS-based tasks in \bench). We encourage researchers to develop new methods that can utilize both satellite imagery and street-level imagery, where available.

\subsection{SDG 2: Zero Hunger}

\subsubsection{Cropland mapping}

Following \citealt{wang2020weakly}, this task evaluates the model's performance on semantic segmentation. The goal for the task with a single pixel label is to predict whether the single labeled pixel in the image is cropland or not. The goal for the task with image-level labels is to detect whether the majority ($\ge$50\%) of pixels in an image are classified to the cropland category. In both cases, the model is a U-Net trained using the binary cross entropy loss defined as
\begin{equation}
    l(y,\hat{y})=-[y \log \hat{y} + (1 - y)\log(1- \hat{y})],
\end{equation}
where $y$ is either the single-pixel label or the image-level binary label and $\hat{y}$ is the single-pixel or image-level model prediction. The evaluation metric is test set accuracy, precision, recall, and F1 scores. Details about the dataset are provided in \Cref{app:sec:crop_mapping}.

\paragraph{Comparison with Related Works}

As mentioned in \Cref{app:sec:crop_mapping}, existing cropland products have been created using SVMs or tree-based algorithms that take into account a single pixel at a time \cite{buchhorn2020copernicus, friedl2002global, xiong2017automated}. In Togo, \citealt{kerner2020rapid} showed that a multi-headed LSTM (still trained on single pixels) outperformed these classifiers on cropland prediction. Since \bench{}'s cropland dataset is a static mosaic over the growing season, we chose to stick with the U-Net in \citealt{wang2020weakly} as the backbone architecture for the baseline. Segmentation models that are more state-of-the-art than the U-Net would be good candidates to surpass this baseline. Active learning or semi-supervised learning methods could also beat a baseline that uses randomly sampled weak labels for supervision. Future updates to this cropland dataset can include the temporal dimension for cropland mapping as well.

\subsubsection{Crop type mapping in Ghana and South Sudan}
The architecture described in \citealt{rustowicz2019semantic} obtained an average F1 score and overall accuracy of 0.57 and 0.61 in Ghana and 0.70 and 0.85 in South Sudan respectively, demonstrating the difficulty of this task.
We use the same train, validation, and test splits as \cite{rustowicz2019semantic}. However, we use the full $64\times64$ imagery provided, while \cite{rustowicz2019semantic} further subdivided imagery into $32\times 32$ pixel grids due to memory constraints. We also include variable-length time series with zero padding and masking, while \cite{rustowicz2019semantic} trimmed the respective time series down to the same length. We include variable-length time series with the reasoning that future research should be extendable to variable length time-series imagery. Due to these changes, we do not include baseline models from \cite{rustowicz2019semantic} for this iteration of the dataset. We provide more details in \Cref{app:sec:croptype1}.

\paragraph{Comparison with Related Works}

Like cropland maps, most operational works classifying crop types employ SVM or random forest classifiers \cite{cdl,inglada2015assessment}. The baseline model that we use from \citealt{rustowicz2019semantic} improves upon these by using an LSTM-CNN. Recent models used in other, non-operational works include 1D CNNs and 3D CNNs \cite{wang2020mapping} and kNN \cite{kerner2020fieldlevel}. A review from this year comparing five deep learning models found that 1D CNN, LSTM-CNN, and GRU-CNN all achieved high accuracy on  classifying crop types in China, with differences between them statistically insignificant \cite{zhao2021evaluation}.

\subsubsection{Crop type mapping in Kenya}

The crop type data in Kenya come from three regions: Bungoma, Busia, and Siaya. We provide ML researchers with the option of splitting fields randomly or by region. The former setup would test the crop type classifier's ability to distinguish crop type in-domain, while the latter would test the classifier's out-of-domain generalization. In \Cref{tab:benchmark}, we show results for the latter from \cite{kluger2021two}.

In \citealt{kluger2021two}, the authors trained on one region and tested on the other two in order to design algorithms that transfer from one region to another. In order to generalize across regions, they corrected for (1) crop type class distribution shifts and (2) feature shift between regions by estimating the shift using a linear model. The features used are the coefficients of a harmonic regression on Sentinel-2 time series. (In the field of remote sensing, the Fourier transform is a common way to extract features from time series \cite{jin2019smallholder}.) The results from \citealt{kluger2021two} show that harmonic features achieve a macro F1-score of 0.30 when averaged across the three test sets, highlighting the difficulty of this problem. Note that this baseline did not include the Non-crop class in the analysis.

\paragraph{Comparison with Related Works}

We expect that, for in-domain crop type classification, methods mentioned previously (1D CNN, LSTM-CNN, GRU-CNN) will outperform the random forests and LDA used in \cite{jin2019smallholder} and \cite{kluger2021two}. However, for cross-region crop type classification, \citealt{kluger2021two} found that a simpler LDA classifier outperformed a more complex random forest. Nonetheless, deep learning-based algorithms that are designed for out-of-domain generalization could outperform the baseline. To our knowledge, these methods have not yet been tested on crop type mapping.

\subsubsection{Crop yield prediction}

The task is to predict the county-level crop yield for that season, in metric tonnes per cultivated hectare, from the MODIS spectral histograms. We split the task into three separate subtasks of crop yield prediction in the United States, Argentina, and Brazil, and provide a 60-20-20 train-validation-test split. For each subtask, we encourage the usage of transfer learning and other cross-dataset training, especially due to the imbalance in data availability, between the United States, Argentina, and Brazil.
Averaged across the years from 2012--2016, the benchmark models in \citealt{wang2018transfer} achieve an RMSE of 0.62 trained and evaluated on Argentina, 0.42 trained and evaluated on Brazil, and 0.38 using transfer learning on an Argentina-trained model to evaluate on Brazil.
Averaged across 2011--2015, the benchmark models in \citealt{you2017deep} achieve an RMSE of 0.37 trained and evaluated on the United States.
However, we note that our datasets and splits are not identical to the original papers, so the results are not directly transferable.

\paragraph{Comparison with Related Works}

Several past works apply machine learning algorithms to human-engineered satellite features such as linear regression over NDVI \cite{quarmby1993ndvi} and EVI2 \cite{bolton2013forecasting}. The papers that originally compiled \bench{}'s datasets compared against these methods and outperformed them. A few other works, like \citealt{sun2019soybean}, apply different architectures to spectral histograms similar to those provided in \bench. Still other methods report results trained on ground-based data, such as ground-level images of crops \cite{tedesco2020cotton}, but these datasets have not been made public.

\subsubsection{Farmland parcel delineation}
Given an input satellite image, the goal is to output the delineated boundaries between farm parcels, or the segmentation masks of farm parcels~\cite{aung2020farm}. Similar to \cite{aung2020farm}, given the predicted delineated boundaries of an image, we use the Dice score as the evaluation metric
\begin{equation}
\label{eq:dice_score}
    DICE=\frac{2TP}{2TP+FP+FN},
\end{equation}
where ``TP'' denotes True Positive, ``FP'' denotes False Positive, and ``FN'' denotes False Negative. As discussed in \cite{aung2020farm}, the Dice score~\Cref{eq:dice_score} has been widely used in image segmentation tasks and is often argued to be a better metric than accuracy when class imbalance between boundary and non-boundary pixels exists.

\paragraph{Comparison with Related Works}

While the original paper that compiled \bench{}'s field delineation dataset achieved a Dice score of 0.61 with a standard U-Net \cite{aung2020farm}, we applied a new attention-based CNN developed specifically for field delineation \cite{waldner2021detect} and achieved a 0.87 Dice score. To our knowledge, this is the state-of-the-art deep learning model for field delineation.

\subsection{SDG 13: Climate Action}

The task is binary classification on satellite imagery, where class 0 "no kiln" means there is no brick kiln present in the image and class 1 "yes kiln" means there is a brick kiln. The training-validation split of the provided Sentinel-2 imagery is 80-20. The ResNet50~\cite{he2016deep} model trained in \cite{lee2021scalable} achieved 94.2\% accuracy on classifying high-resolution (1m/pixel) imagery; the authors hand-validated all positive predictions and 25\% of negative predictions. The imagery was not released publicly because it was proprietary, so we report a baseline validation accuracy of 94.5\%, training a ResNet50 model on lower-res Sentinel-2 imagery using only the Red, Blue, and Green bands (B4, B3, B2). In addition to accuracy on the validation set, AUC, precision, and recall are also valuable metrics given the class skew toward negative examples.

\subsection{SDG 15: Life on Land}

\subsubsection{Representation learning for land cover classification}

\citealt{jean2019tile2vec} performed land cover classification using features learned through an unsupervised, contrastive loss algorithm named Tile2Vec. Since the features are learned in entirely unsupervised ways, they can be used with any number of labels to train a classifier. At $n=1000$, Tile2Vec features with a multi-layer perceptron (MLP) classifier achieved 0.55 accuracy; at $n=10,000$, Tile2Vec features with an MLP achieved 0.58 accuracy. Notable also is that Tile2Vec features outperformed end-to-end training with a CNN sharing the same architecture as the feature encoder up to $n=50,000$ labels.

\paragraph{Comparison with Related Works}

\citealt{jean2019tile2vec} was the first to apply the distributional hypothesis from NLP to satellite imagery in order to learn features in an unsupervised way. Tile2Vec features outperformed features learned via other unsupervised algorithms like autoencoders and PCA. Methods that have not yet been tried but could yield high-quality representations include inpainting missing tiles, solving a jigsaw puzzle of scrambled satellite tiles, colorization, and other self-supervised learning techniques. Recently, \cite{rolf2021a} proposed a representation learning approach that uses randomly sampled patches from satellite imagery as convolutional filters in a CNN encoder, which could also be tested on this dataset.

\subsubsection{Out-of-domain land cover classification}

\citealt{wang2020meta} defined 1-shot, 2-way land cover classification tasks in each region, and compared the performance of a meta-learned CNN with pre-training/fine-tuning and training from scratch. The meta-learned CNN performed the best on the meta-test set. The meta-learning algorithm used was model-agnostic meta-learning (MAML). The MAML-trained model achieved an accuracy of 0.74, F1-score of 0.72, and kappa score of 0.32 when averaged over all regions in Sub-Saharan Africa in the meta-test set. Unlike other classification benchmarks in \bench, this benchmark uses the kappa statistic to evaluate models because accuracy and F1-scores can vary widely across regions depending on the class distribution, and it is not clear whether an accuracy or F1-score is good or bad from the values alone.

We note that, as previously mentioned, existing land cover products tend to be less accurate in low-income regions such as Sub-Saharan Africa than in high-income regions. As a result, the MODIS land cover product used as ground truth will have errors in low-income regions. We suggest users also apply meta-learning and other transfer learning algorithms using other continents (\emph{e.g.}, North America, Europe) as the meta-test set for algorithm evaluation purposes.

\paragraph{Comparison with Related Works}

To our knowledge, \cite{wang2020meta} and \cite{russwurm2020meta} (same authors) were the first works to apply meta-learning to land cover classification in order to simulate sharing knowledge from high-income regions to low-income ones. The baseline cited in \Cref{tab:benchmark} uses MAML, which is one of the most widely-used meta-learning algorithms. As the field of meta-learning is advancing quickly, we hope ML researchers will evaluate the latest meta-learning algorithms on this land cover classification dataset.

\section{Ethical Concerns}
\label{app:sec:ethics}

Because the SDGs are high stakes issues with direct societal impacts ranging from local to global levels, it is imperative to exercise caution in addressing them. Researchers must be aware of and work to address the potential biases in the training data and in the generated predictions.
For example, current models have been observed to over-predict wealth in poor regions and under-predict wealth in rich regions \cite{jean2016combining}. If such a model were used to distribute aid, the poor would receive less than they should.
Much work remains to be done to understand and rectify the biases present in ML model predictions before they can play a significant role in policy-making.

Because the \bench{} dataset involves remote sensing and geospatial data that covers areas with private property, data privacy can be a concern.  We summarize below the risks of revealing information about individuals present in each dataset.
\begin{itemize}
\item For our survey data (see \Cref{tab:dhs_surveys,tab:lsms_surveys}), the geocoordinates for DHS and LSMS survey data are jittered randomly up to 2km for urban clusters and 10km for rural clusters to protect survey participant privacy \cite{burgert2013geographic}. Furthermore, geocoordinates and labels are only released for ``clusters'' (roughly villages or small towns); no household or individually identifiable data is released.
\item Mapillary images, as well as satellite images from Landsat, Sentinel-1, Sentinel-2, MODIS, DMSP, NAIP, and PlanetScope, are all publicly available. In particular, all of these satellites other than PlanetScope are low-resolution. Mapillary automatically blurs faces of human subjects and license plates, it allows users who upload images to manually blur parts of images for privacy. Thus it is very difficult to get individually identifiable information from these images, and we believe that they do not directly constitute a privacy concern.
\item The crop yield statistics, made publicly available by the governments of the US, Argentina, and Brazil, are published after aggregating over such large areas that the yields of individual farms cannot be derived.
\item The crop type dataset released by \citealt{rustowicz2019semantic} has no geolocation information that would allow tracing to individuals. The satellite imagery released also has noise added so that it is more difficult to identify the original location and time that the imagery was taken. The crop type dataset released in Kenya likewise does not include geolocation.
\item For the field delineation dataset, boundary shapefiles are publicly available from the French government as part of the European Union's Common Agricultural Policy \cite{aung2020farm}. The data has been stripped of any identifying information about farmers.
\item Brick kilns labels were generated by one of the authors under the guidance of domain experts. The version of this dataset released in \bench{} consists of Sentinel-2 imagery, from which very few privacy-concerning details can be seen (see \Cref{app:fig:brick_kiln}).
\item The labels used for the representation learning task and out-of-domain land cover classification task are products of other machine learning algorithms. They are publicly available and do not reveal information about individuals.
\end{itemize}

\end{document}